\documentclass[sn-basic,Numbered]{sn-jnl}

\usepackage{graphicx}%
\usepackage{multirow}%
\usepackage{amsmath,amssymb,amsfonts}%
\usepackage{amsthm}%
\usepackage{mathrsfs}%
\usepackage[title]{appendix}%
\usepackage{xcolor}%
\usepackage{textcomp}%
\usepackage{manyfoot}%
\usepackage{booktabs}%

\usepackage{amssymb}
\usepackage{xcolor}
\usepackage{hyperref}

\usepackage[switch]{lineno}
\usepackage{multirow,multicol}
\usepackage{colortbl}
\usepackage{pgfplots}
\usepackage{pgfplotstable}

\theoremstyle{thmstyleone}%
\theoremstyle{thmstyletwo}%

\theoremstyle{thmstylethree}%

\begin{document}

\title[CrossPT: Exploring Cross-Task Transferability through Multi-Task Prompt Tuninge]{CrossPT: Exploring Cross-Task Transferability through Multi-Task Prompt Tuning}


\author*[1,2]{\fnm{Ahmad} \sur{Pouramini}}\email{ahmad.pouramini@ut.ac.ir}

\author[1]{\fnm{Hesham} \sur{Faili}}\email{hfaili@ut.ac.ir}
\equalcont{These authors contributed equally to this work.}

\affil*[1]{\orgdiv{School
		of Electrical and Computer Engineering, College of
		Engineering}, \orgname{University of Tehran}, \orgaddress{\street{North Kargar Street}, \city{Tehran}, \postcode{515-14395}, \state{Tehran}, \country{Iran}}}

\affil[2]{\orgdiv{Department of Computer Engineering}, \orgname{Sirjan University of Technology}, \orgaddress{\city{Sirjan}, \postcode{78137-33385}, \country{Iran}}}

\abstract{
	Prompt tuning offers a parameter-efficient way to adapt large pre-trained language models to new tasks, but most existing approaches are designed for single-task settings, failing to share knowledge across related tasks. We propose Cross-task Prompt Tuning (CrossPT), a modular framework for multi-task prompt tuning that enables controlled knowledge transfer while maintaining task-specific specialization. CrossPT decomposes each target prompt into shared, pre-trained source prompts and task-specific private prompts, combined via a learned attention mechanism. To support robust transfer, we systematically investigate key design factors including prompt initialization, balancing shared and private prompts, number of source prompts, learning rates, task prefixes, and label semantics. Empirical results on GLUE and related benchmarks show that CrossPT achieves higher accuracy and robustness compared to traditional prompt tuning and related methods, particularly in low-resource scenarios, while maintaining strong parameter efficiency.
}


\maketitle

\section{Introduction}

Modern NLP systems increasingly rely on parameter-efficient adaptation methods to customize large pre-trained language models (PLMs) for new tasks without updating all model parameters. Among these, \emph{prompt tuning} has become popular for its simplicity and low memory footprint: it learns a small set of continuous prompt embeddings while keeping the PLM frozen \cite{lester2021power}. This allows adaptation to many tasks with minimal added parameters for both language models \cite{peft} and vision-language models \cite{visualprompt}.

Despite this efficiency, most existing prompt tuning methods are designed for \emph{single-task learning}, where prompts are optimized independently for each task without sharing knowledge across them. This design is limiting in multi-task settings, where tasks often share semantic structure, labels, or data domains that can be exploited for transfer. Learning isolated prompts from scratch for each task misses opportunities for cross-task knowledge sharing---especially problematic in few-shot scenarios where data is scarce. Moreover, with only a single prompt vector per task, it becomes difficult to represent complex task relationships or adapt to varying label spaces, risking both underfitting of shared patterns and overfitting of task-specific noise.

These challenges motivate the need for a more \emph{modular and transferable} approach to prompt tuning that can share generalizable representations across tasks while preserving task-specific specialization in a parameter-efficient manner.

\textbf{Problem Definition and Challenges.} We focus on the problem of \emph{multi-task prompt tuning} for NLP, where the goal is to efficiently adapt a single pre-trained model to multiple related tasks. The key challenges include:
\begin{itemize}
	\item \textbf{Balancing generalization and specialization}: How to capture shared knowledge across tasks while maintaining task-specific performance.
	\item \textbf{Efficient parameter usage}: How to achieve this balance with minimal additional parameters and compute.
	\item \textbf{Transfer in few-shot scenarios}: How to improve low-resource task performance by leveraging high-resource related tasks.
\end{itemize}

\textbf{Our Approach.} In this paper, we introduce \emph{Cross-task Prompt Tuning} (CrossPT), a modular, parameter-efficient framework designed to address these challenges. CrossPT decomposes each target task's prompt into two components: \emph{shared source prompts} learned on individual source tasks, and a \emph{task-specific private prompt}. An attention mechanism dynamically combines these components to create the final prompt for each task. This design enables adaptive knowledge transfer while maintaining task-specific specialization.

CrossPT supports flexible configurations to study how the balance between shared and private prompts affects performance, especially in few-shot settings. Our experiments on GLUE and related benchmarks demonstrate significant improvements over conventional prompt tuning and strong parameter efficiency compared to full fine-tuning.

\paragraph{Our key contributions are as follows:}
\begin{itemize}
	\item We propose CrossPT, a modular framework for multi-task prompt tuning that explicitly decomposes task prompts into shared source prompts and task-specific private prompts to enable controlled knowledge transfer.
	\item We design an attention mechanism to dynamically weight source and private prompts for each task, supporting task-adaptive specialization while reusing shared representations.
	\item We systematically study the impact of critical design choices, including the number of source prompts, prompt initialization strategies, learning-rate balancing, and task label design, offering practical guidance for effective multi-task prompt tuning.
		\item We introduce new metrics (e.g., Weighted Similarity to Source Prompts) to quantitatively analyze knowledge sharing and specialization within the framework.
	\item We empirically validate CrossPT on the GLUE tasks, demonstrating improved accuracy, robustness, and transferability, especially in few-shot settings.
\end{itemize}

In the remainder of this paper, we review related work, describe the CrossPT framework in detail, analyze design choices, present extensive experimental results, and discuss implications and limitations of our approach.

\begin{figure}[t]
	\centering
	\includegraphics[width=1\linewidth]{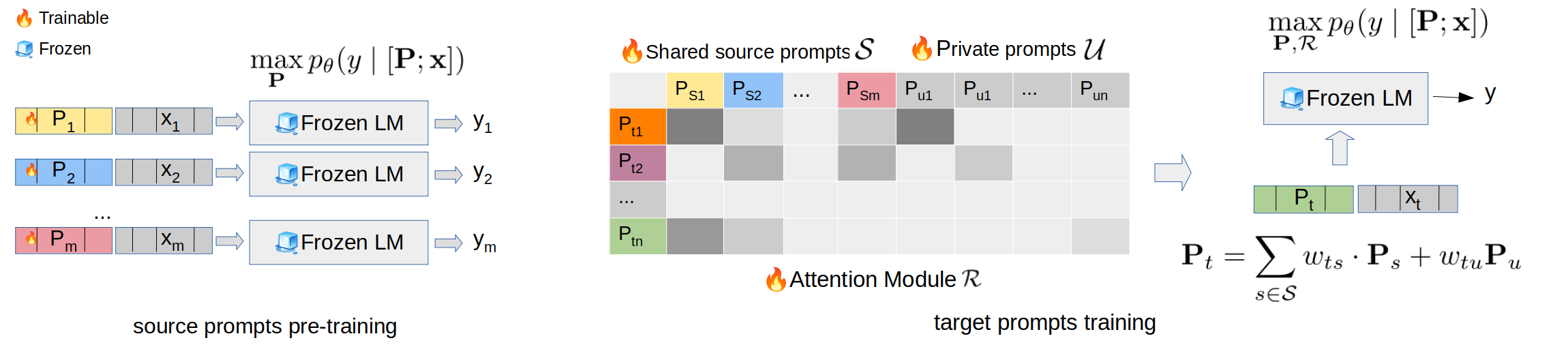}
	\caption{Overview of the CrossPT framework. Stage 1 learns source prompts for individual tasks. Stage 2 combines shared source prompts and task-specific private prompts via learned attention weights (visualized as heatmaps) to construct target task prompts.}
	\label{fig:overal}
\end{figure}

\section{Method}
Figure~\ref{fig:overal} illustrates the CrossPT framework, which consists of two main stages: source prompt pre-training and target prompt training.
In the first stage, a set of source prompts $[\mathbf{P}_1, \ldots, \mathbf{P}_m]$ is pre-trained on multiple tasks, including both the target tasks and other relevant tasks.

In the second stage, the learned source prompts are used to learn the soft prompt $\mathbf{P}_t$ for each target task $t$. This stage can be performed separately for each target task or jointly with multi-task learning across all tasks.
We mainly evaluate CrossPT in multi-task settings. Learning a shared prompt space in multi-task settings can be practically challenging, as it requires capturing commonalities across different source tasks while minimizing interference. In the following sections, we describe the details of each stage.

\subsection{Source Prompt Training}
To initialize the multi-task training phase, we first obtain source prompts $\mathbf{P}_1, \ldots, \mathbf{P}_m$ for the source tasks via prompt tuning. Prompt tuning adapts a pre-trained language model $\mathcal{LM}$ to a specific task by prepending a small sequence of trainable prompt embeddings to the input, while keeping all other model parameters frozen.

We define a template of the form
\[
\mathcal{T} = \{p_1, \ldots, p_k, x_1, \ldots, x_l\}
\]
where $p_1, \ldots, p_k$ are prompt tokens and $x_1, \ldots, x_l$ represent the input sequence. Each prompt token $p_i$ is mapped to a continuous embedding $\mathbf{e}_i \in \mathbb{R}^d$, and the soft prompt is denoted by
\[
\mathbf{P} = [\mathbf{h}_1, \ldots, \mathbf{h}_k] \in \mathbb{R}^{k \times d}
\]
where $\mathbf{h}_i$ is the transformed embedding of the $i$-th prompt token, and $d$ is the embedding dimension of the language model.

To obtain $\mathbf{P}$, the model optimizes the following objective:
\[
\max_{\mathbf{P}} \; p_\theta(\mathbf{y} \mid [\mathbf{P}; \mathbf{x}])
\]
where $\mathbf{x} = [x_1, \ldots, x_l]$ is the input sequence, $\mathbf{y}$ is the target output, and $[\mathbf{P}; \mathbf{x}]$ denotes their concatenation as input to $\mathcal{LM}$. Only the prompt embeddings $\mathbf{P}$ are updated during training; the language model $\mathcal{LM}$ remains frozen.

Directly optimizing $\mathbf{P}$, however, can lead to instability and poor local minima \cite{prefix, gptund}. To alleviate this, a lightweight neural network---called a prompt encoder---is used to generate the prompt embeddings from a learned set of input tokens. In the simplest and most effective variant, this encoder reduces to a single linear transformation:
\[
\mathbf{h}_i = \mathbf{W} \mathbf{e}_i + \mathbf{b}
\]
where $\mathbf{e}_i$ is the raw embedding of the $i$-th prompt token, and $\mathbf{W} \in \mathbb{R}^{d \times d}$ and $\mathbf{b} \in \mathbb{R}^{d}$ are trainable parameters.

This formulation offers a favorable balance between performance and efficiency, with minimal added parameters. We also explored deeper prompt encoders---such as MLPs and LSTM-based networks as proposed in \cite{gptund}---but found that the simple linear projection achieves comparable or better performance with significantly lower complexity.

\subsection{Target Prompt Training}

After acquiring the source prompts, the next stage of CrossPT focuses on training target prompts for new tasks. For each target task, the target prompt is constructed as a weighted combination of shared \textit{source} prompts and a \textit{private} task-specific prompt.

Let $\mathcal{S}$ be the index set of the source prompts, and let $u$ denote the index of the private prompt for the target task $t$. The target prompt $\mathbf{P}_t$ is computed as:

\begin{align}
	\mathbf{P}_t = \sum_{s \in \mathcal{S}} w_{ts} \cdot \mathbf{P}_s + w_{tu} \cdot \mathbf{P}_u,
	\label{eq:pt}
\end{align}

where $\mathbf{P}_s$ is the soft prompt from source task $s$, shared across all target tasks, and $\mathbf{P}_u$ is the task-specific private prompt for target task $t$. The combination weights $w_{ts}$ (for sources) and $w_{tu}$ (for the private prompt) are computed via an attention module $\mathcal{R}$. This module learns to assign higher weights to source prompts that are most relevant to the target task while also maintaining the private prompt's capacity to capture task-specific nuances.

When the source prompts provide limited relevant information, the attention module can reduce their influence, relying more heavily on the private prompt $\mathbf{P}_u$ to ensure effective adaptation. By using a shared attention module $\mathcal{R}$ that learns these weights jointly for all target tasks, the model balances shared and private information in a principled way. 

The weights $w_{ts}$ and $w_{tu}$ are model parameters generated as logits by the attention module. These logits are then normalized via softmax to produce a valid convex combination of source and private prompts:

\[
\mathbf{w}_t = \text{softmax}(\mathbf{z}_t),
\]

where $\mathbf{z}_t$ are the raw scores produced by $\mathcal{R}$ for task $t$. These weights are learned jointly with the private prompt $\mathbf{P}_u$ through backpropagation.

\paragraph{Normalization.}
To better reflect task relevance, we use an adaptive temperature in the softmax normalization. As the number of source prompts $M$ increases, standard softmax tends to yield flatter distributions, making it harder to select truly relevant prompts. To counteract this, we define the temperature parameter as

\[
\tau = \frac{1}{M},
\]

and compute the weights as

\[
w_i = \frac{\exp\left(z_i / \tau \right)}{\sum_{j} \exp\left(z_j / \tau \right)}.
\]

This design ensures that as $M$ grows, the temperature decreases, sharpening the softmax distribution and making the model more selective in assigning weights.

\paragraph{Fast and Slow Learning.}
Training the attention module $\mathcal{R}$ can be challenging due to instability and overfitting \cite{modular}. To address this, we employ different learning rates for different components. A higher learning rate is used for the attention module to enable rapid adaptation by leveraging existing source prompts. In contrast, lower learning rates are assigned to the private and source prompts to ensure stable, gradual learning of their embeddings. This approach follows strategies explored in prior work \cite{rel-attempt}. We also experiment with different learning rates for source versus private prompts to fine-tune their balance during training.

\paragraph{Inference.}
At inference time, the shared source prompts, all private prompts, and the trained attention module $\mathcal{R}$ are loaded once. The attention weights for all target tasks are precomputed and stored in a matrix, where each row corresponds to a target task. For a given target task $t$, the target prompt $\mathbf{P}_t$ is generated by applying the attention weights as in Equation~\ref{eq:pt}. This target prompt is then prepended to the input sequence $\mathbf{x}$ and passed to the language model $\mathcal{LM}$, which generates the output $\mathbf{y}$ by attending jointly to the prompt and the input.

\subsection{Prompt Training Configurations}
\label{sec:design-choices}
Building on the previous section, this part details the different configurations explored to evaluate the impact of various design choices in target prompt training. Specifically, we examine the influence of the following factors:

\begin{itemize}
	\item Presence and initialization of source prompts using pre-trained source prompts.
	\item Learning of source prompts during target prompt training.
	\item Presence and initialization of task-specific (private) prompts using the obtained source prompts.
\end{itemize}

\begin{table*}[th!]
	\centering
	
	\caption{Configurations for training target prompts with and without source prompts. 'S' indicates use of source prompts; 'P' indicates use of private prompts. 'I' denotes initialization (using pre-trained prompts), while 'L' indicates whether the source prompts are learned or kept frozen. Private prompts are always learned.}
	
	\label{table:mets}
	\begin{tabular}{|c|c|c|c|c|c|c|}
		\hline
		& \small Use Source & \small Initialize Source & \small Learn Source  & \small Use Private & \small Initialize Private   \\
		\hline
		P & \texttimes & --- & --- & \checkmark & \texttimes \\
		\hline
		PI & \texttimes & --- & --- & \checkmark & \checkmark \\
		\hline
		SL & \checkmark  & \texttimes  & \checkmark & \texttimes  & ---   \\
		\hline
		SLP & \checkmark & \texttimes  & \checkmark  & \checkmark  & \texttimes  \\
		\hline
		SIL & \checkmark  & \checkmark & \checkmark  & \texttimes & ---  \\
		\hline	
		SIP & \checkmark  & \checkmark  & \texttimes & \checkmark  & \texttimes   \\
		\hline
		SILP & \checkmark  & \checkmark  & \checkmark  & \checkmark & \texttimes   \\
		\hline
	\end{tabular}
\end{table*}

In theory, over 20 possible combinations arise from these factors. However, our experiments focus on the most meaningful configurations, summarized in Table~\ref{table:mets}. Several of these configurations correspond to existing methodologies or standard prompt tuning variants. For example, the case where no source prompts are used and private prompts are learned independently corresponds to standard prompt tuning, denoted as P in the table.

To allow fair comparison with prompt tuning initialized from source prompts, we introduce a variant (PI) where private prompts are initialized with the corresponding pre-trained source prompts from Stage 1. This setup tests whether transferring source prompt knowledge improves target-task adaptation.

Other configurations can be interpreted directly from the table. For instance, SILP refers to a setting where source prompts are initialized using pre-trained prompts, learned further during target prompt training, and combined with a private prompt (always learned). The presence of 'P' in the label indicates the use of a private prompt in the final target prompt composition.

For scenarios like SL or SLP, source prompts are initialized randomly rather than using pre-trained prompts. We also investigate two main sub-cases:

\begin{itemize}
	\item \textbf{SL} and \textbf{SLP}: a single shared source prompt is used for all target tasks.
	\item \textbf{SLN} and \textbf{SLPN}: a dedicated source prompt is allocated for each target task.
\end{itemize}

These variations allow us to analyze the effect of the number of source prompts and their sharing scheme. In the following sections, we present the experimental results for these methods and provide a detailed comparison of their performance across different tasks.

\section{Experiments}
\subsection{Tasks}
We assess the performance of CrossPT across various tasks, including eight tasks from the GLUE benchmark \cite{glue} as first group and and six tasks other tasks as second group \cite{superglue}.  The datasets are publicly available on  Hugging Face's model hub.\footnote{Hugging Face Model Hub. GLUE and SuperGLUE Datasets. Retrieved from \url{https://huggingface.co/datasets/glue}, \url{https://huggingface.co/datasets/superglue}}

in Group 1, the GLUE tasks encompass MNLI, QNLI, RTE (NLI), COLA (grammatical acceptability), SST2 (sentiment analysis), STSB (semantic similarity), MRPC, and QQP (paraphrase detection). 

For Group 2, we include MultiNLI, SciTail (NLI), IMDB,  Yelp-Polarity (sentiment analysis), PAWS, and MRPC (paraphrase detection). We also included STSB2 where the range of similarty numbers where divided to three categories (high, medium and low).

\subsection{Implementation Details}
We use the T5-base language model \cite{t5} as our base model for both source prompt tuning and all subsequent experiments and baselines. The prompt length is fixed at 10 tokens, which are prepended to the input sequence during training.

We employ a linear network as the prompt encoder, which generates a prompt embedding of size $d$ matching the language model's embedding size for each prompt token.

Training uses the Adam optimizer with a batch size of 16. Techniques involving initialization of source or private prompts were trained for 20 epochs, while methods without prompt initialization (SL, P, and SLP) were trained for an additional 10 epochs to ensure fairness. For method comparison, results are reported on 300 randomly selected test samples from each dataset.

If a dataset lacks a public test split, we use its development set or split it into separate development and test sets.

Learning rates are set as follows: for prompt initialization and source prompt learning methods (e.g., SIL, SILP), we use 0.05 for source prompts and 0.02 for private prompts; for other methods (SLP, SL), 0.07 for private prompts and 0.15 for source prompts; and 0.1 for attention weights.

Our code and full configuration details are publicly available at \url{https://github.com/puraminy/CrossPT}

\section{Results}
\subsection{Comparison of Methods}
\begin{figure}[ht!]
	\centering
	\includegraphics[width=0.45\linewidth]{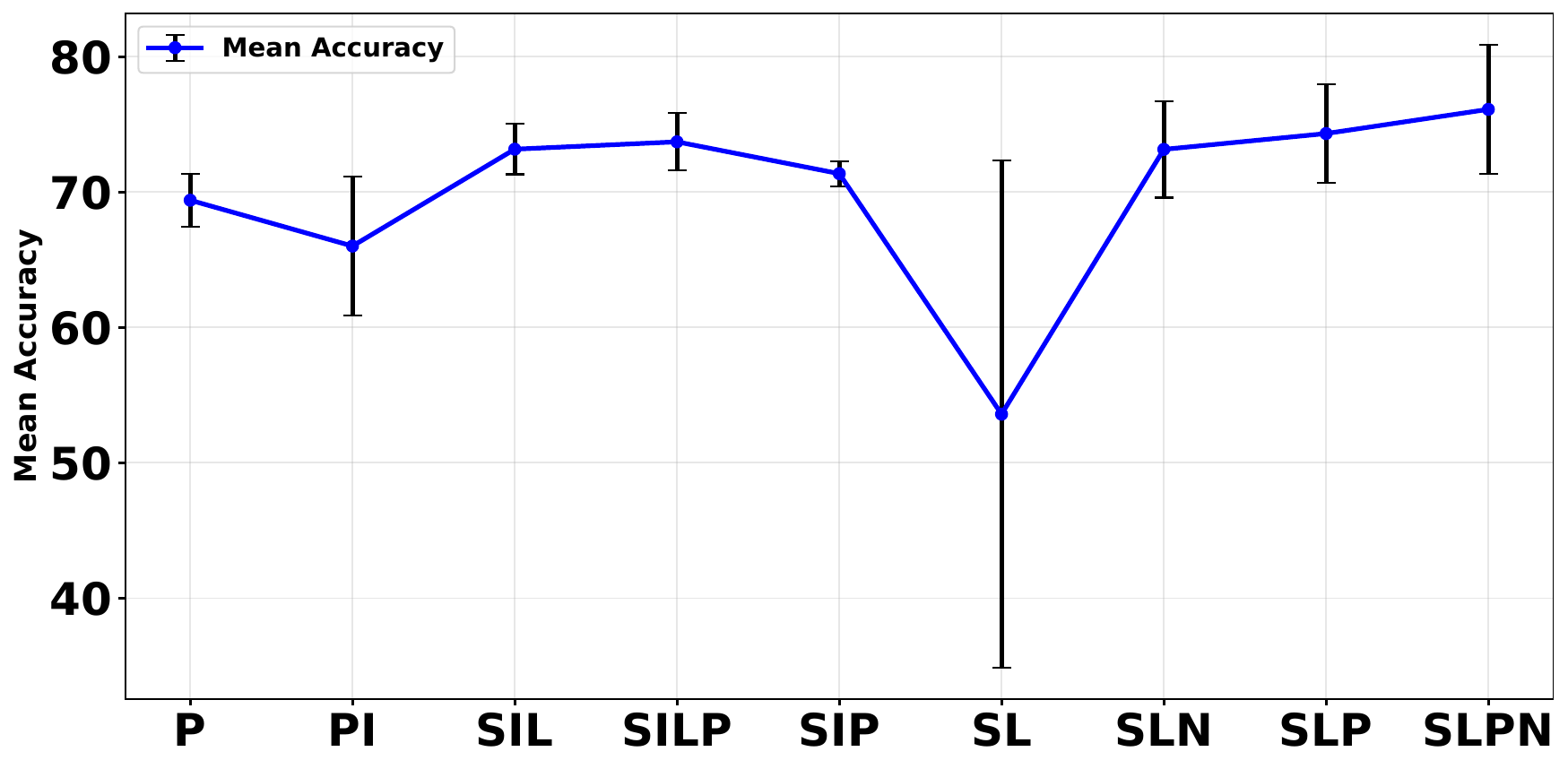}
	\includegraphics[width=0.45\linewidth]{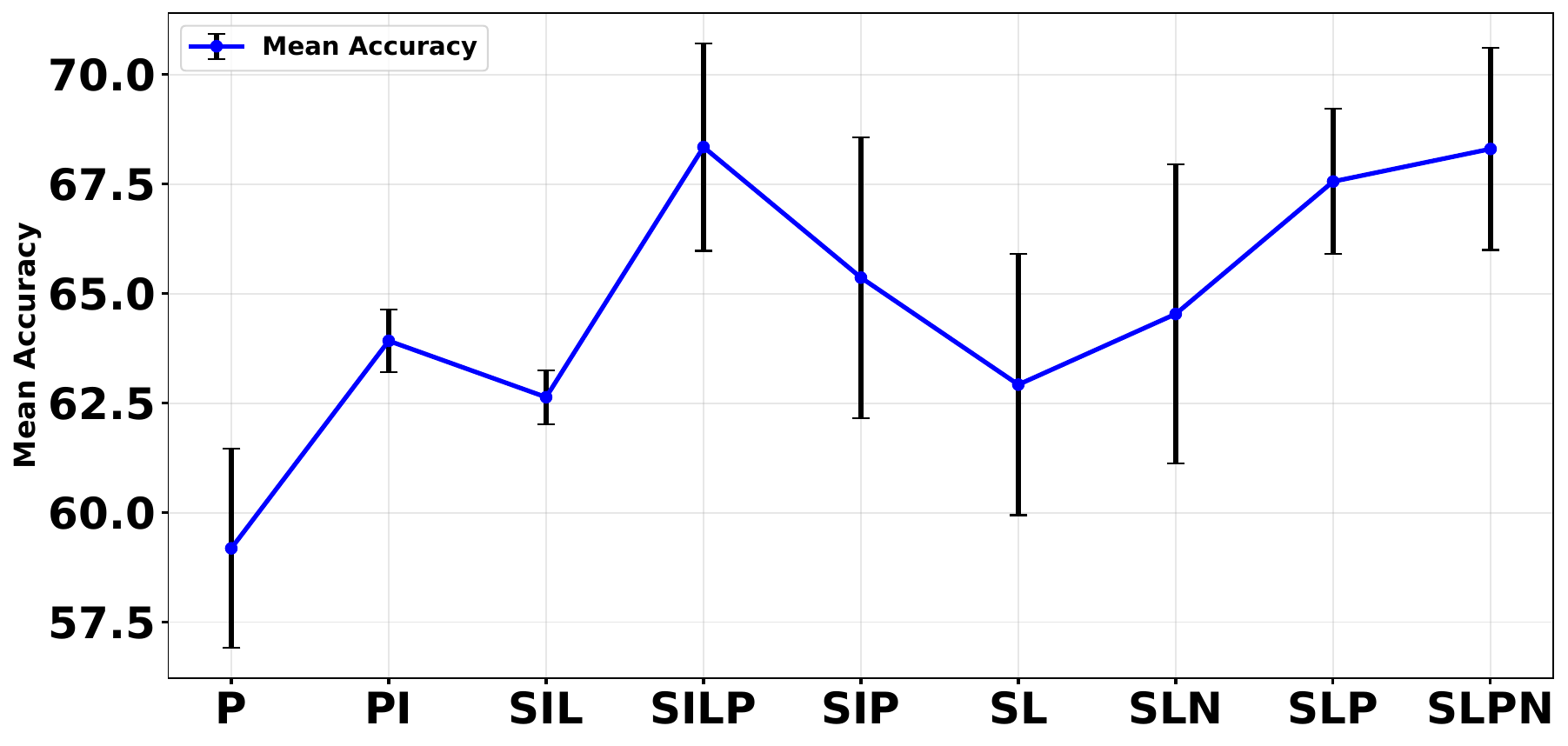}
	\caption{Comparative performance on selected tasks from the GLUE benchmark (top) and Group 2 datasets (bottom) using different proposed methods with 32 training samples. Lines represent mean performance across runs, and shaded areas (or error bars) indicate standard deviation.}
	
	\label{fig:methods}
\end{figure}

Figure \ref{fig:methods} presents the average performance of the methods mentioned in Table \ref{table:mets} on 8 tasks from the GLUE dataset and 6 tasks from Group 2 in a few-shot settings using 32 samples for each task. 
To ensure the reliability of the results,  each experiment was repeated three times with different random seeds. The average performance values of these runs for each task can be found in Appendix A.

\paragraph{Baselines} Among the configurations, P, PI, and SL serve as baselines. Methods P and PI train prompts using only private prompts, while SL learns a single shared prompt across all tasks in a multi-task setting. As observed, the methods that combine one or more source prompts with private prompts---either with initialization (SILP) or without (SLPN)---outperform these baselines. In the case of SL, task interference is high due to the use of a single shared prompt, resulting in lower performance and higher variance. This issue is alleviated by employing multiple source prompts (SIL, SLN) or incorporating private prompts (SLP, SLPN). The difference between using initialization and not using it is marginal; however, SLPN performs slightly better on GLUE, as will be discussed later.

\subsection{Knowledge Transfer between Related Tasks} \label{sec:mrpc-qqp}

\begin{figure}
	\centering
	\includegraphics[width=0.7\linewidth]{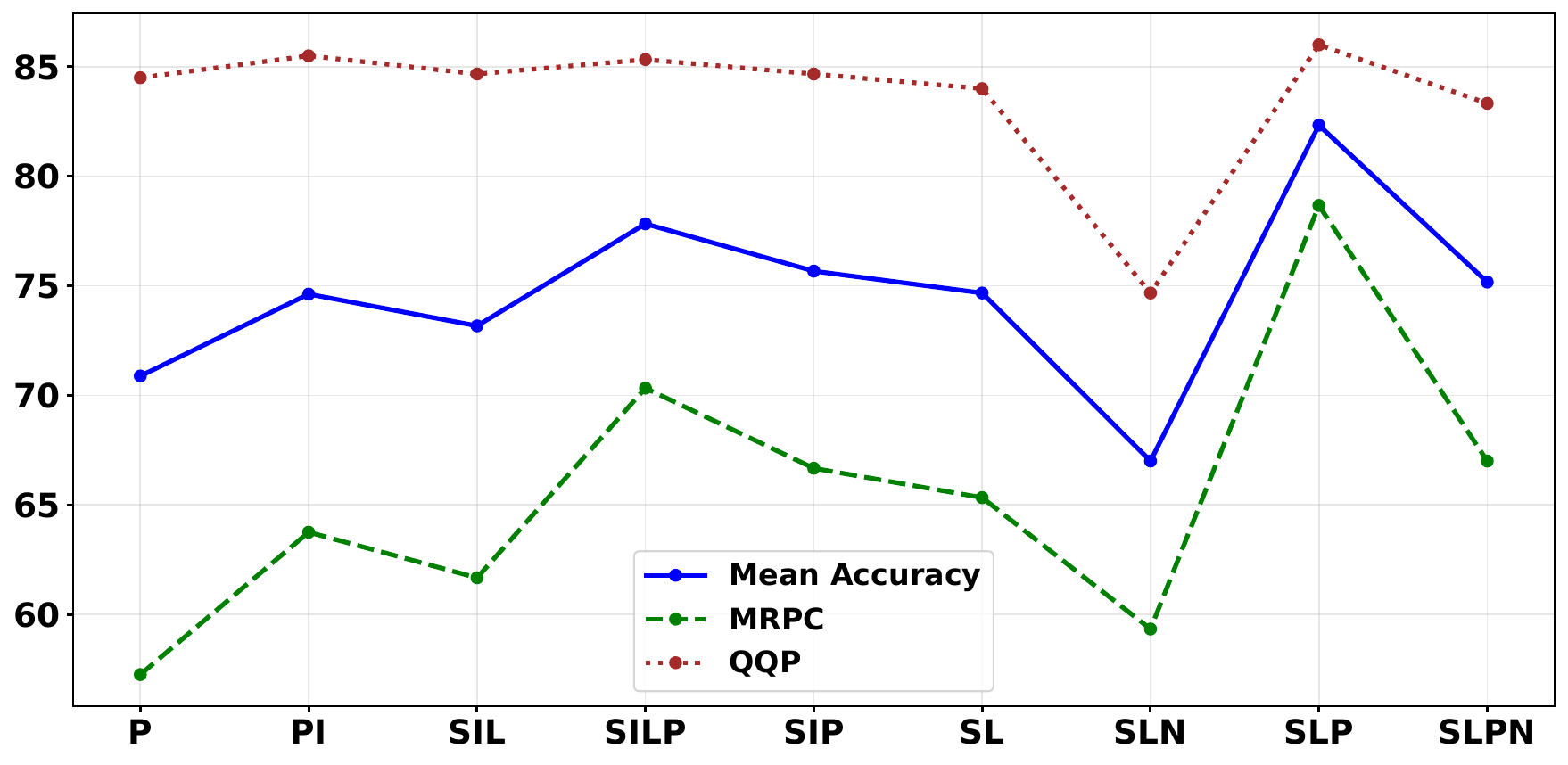}
\caption{Accuracy on QQP and MRPC tasks, along with their mean, across different prompting configurations with 48 training samples. Results illustrate effective knowledge transfer between related tasks when combining shared and task-specific prompts.}

	\label{fig:qqp_mrpc}
\end{figure}

To examine knowledge transfer within our prompt-tuning framework, we analyze performance on two related paraphrase detection tasks, QQP and MRPC. Figure~\ref{fig:qqp_mrpc} plots accuracy for QQP, MRPC, and their mean across all evaluated methods.

Overall, all multi-prompt configurations outperform private-only or source-only baselines, demonstrating the benefit of combining shared and task-specific representations. QQP achieves consistently higher accuracy across methods, suggesting it provides clearer or more learnable paraphrase signals that can support transfer. Importantly, methods that effectively balance shared and private knowledge, such as SLP and SILP, show substantial improvements on MRPC as well. In particular, SLP achieves strong results by using a single shared source prompt to encode common knowledge across tasks while relying on private prompts to resolve conflicts seen in the purely shared SL configuration.

By contrast, SLN---which uses two shared prompts without private prompts---shows the lowest performance. This suggests that splitting shared capacity without task-specific adaptation can fragment general knowledge and amplify conflicts between tasks. Overall, these results highlight that carefully balancing the number of shared prompts and the inclusion of private prompts is crucial for enabling effective cross-task knowledge transfer.

\begin{figure}[ht!]
	\centering
	\includegraphics[width=0.7\linewidth]{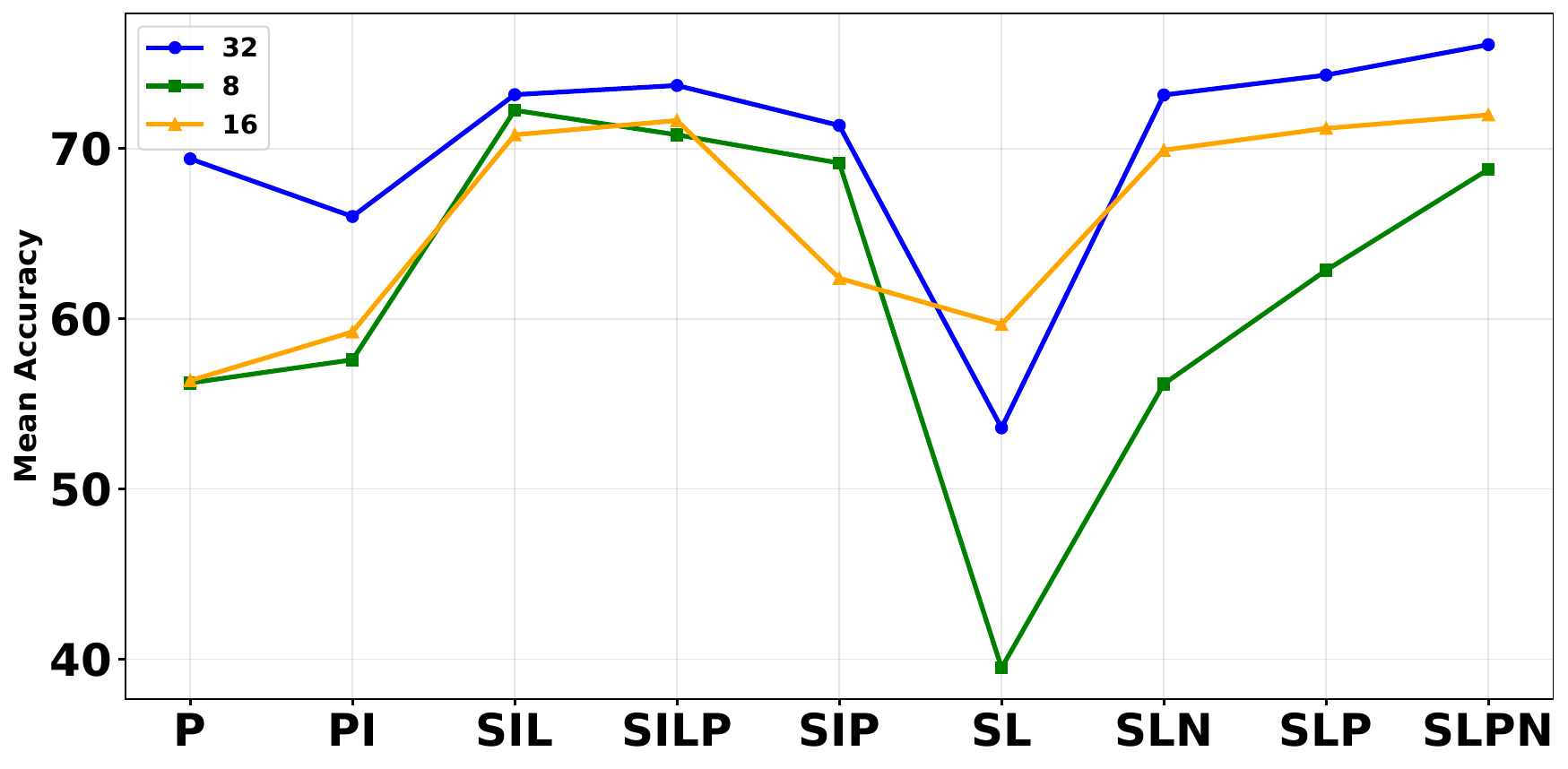}	
	\caption{Few-shot performance on GLUE tasks using different methods and sample sizes (8,16,32 shots). }	
	\label{fig:samples}
\end{figure}

\subsection{Effect of Sample Size}
Figure~\ref{fig:samples} illustrates the performance of various methods across different sample sizes. When data is highly limited (e.g., 8 samples), methods that incorporate initialization (SIL, SIP, and SILP) consistently outperform others, including prompt tuning with initialization (PI). However, as the number of training samples increases, the benefits of initialization diminish. In these cases, methods trained from scratch (SLN, SLP and SLPN), which leverage joint learning across tasks, begin to outperform the initialized approaches. This suggests that prompt initialization is particularly beneficial in very low-resource settings, whereas multi-task learning from scratch becomes more effective as more data becomes available.

\subsection{Effect of Task Prefixes}
In multi-task learning, task prefixes serve as constant indicators (e.g., QNLI, QQP) that can be prepended to the input of each task to help the model distinguish among them. Without such prefixes, inputs for certain tasks such as QQP, QNLI, and STSB all consist of two sentences, making it harder for the model to identify the task type. This ambiguity can affect how source and private prompts encode and share knowledge.

\begin{figure}
	\centering
	\includegraphics[width=0.7\linewidth]{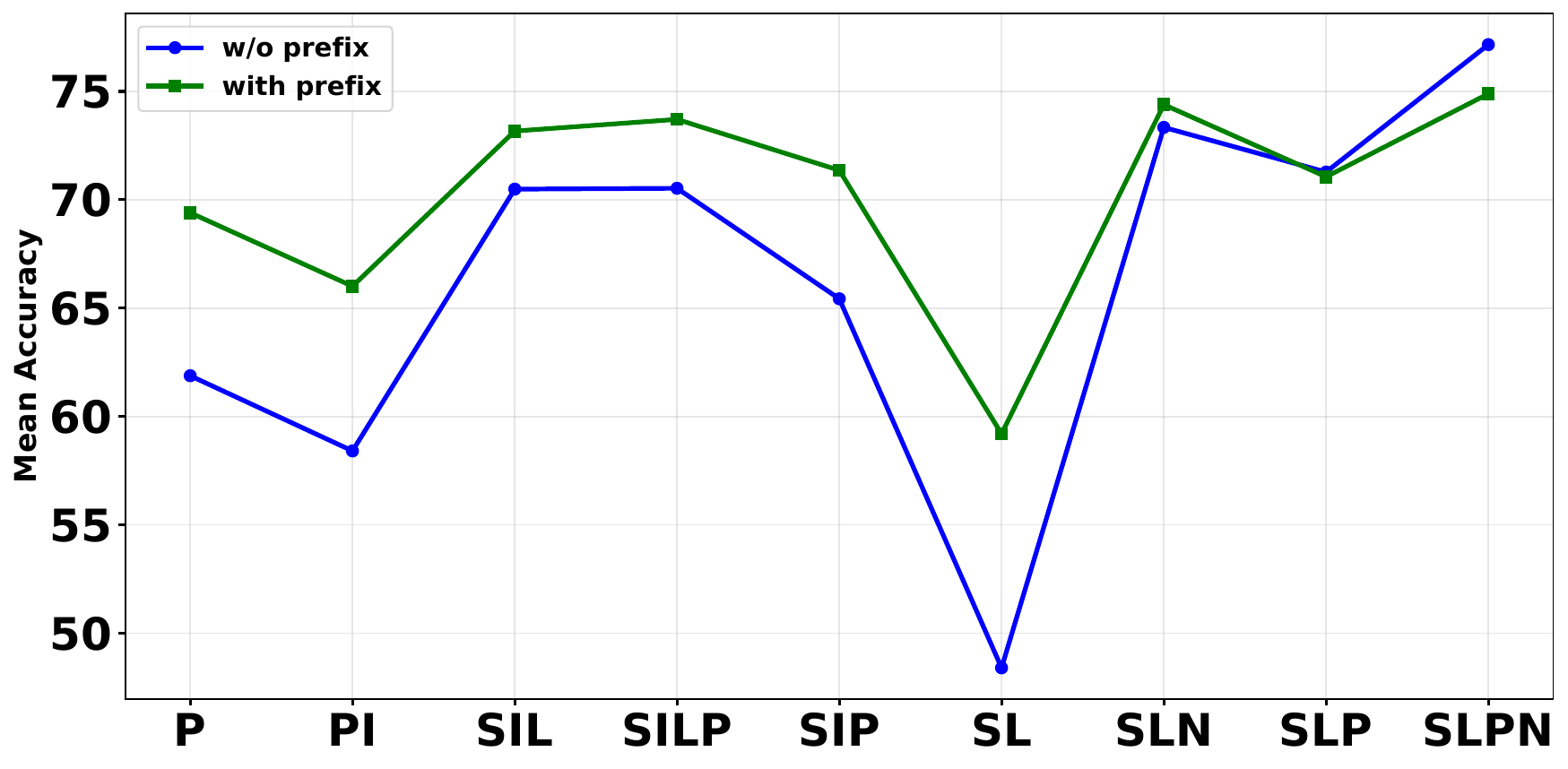}
	\caption{Effect of task prefixes on performance across GLUE tasks using 32 training samples. The figure compares model configurations with and without task-specific prefixes, illustrating that the inclusion of prefixes generally improves performance by providing explicit task cues. }
	
	\label{fig:addp}
\end{figure}

Figure~\ref{fig:addp} illustrates the performance differences observed with and without task prefixes. As expected, including prefixes generally leads to higher performance across different configurations. In particular, for the SL setting, which uses a single shared prompt, the absence of a prefix results in a notable performance drop because the model lacks an explicit mechanism to differentiate tasks. When multiple source prompts or private prompts are used, the model can rely on these prompts to help identify tasks, partially mitigating the effect of missing prefixes. 

The presence of prefixes, particularly in source-only settings such as SIL and SL, can enable better reuse of shared prompts across different tasks by providing explicit task identifiers. However, when private prompts are included, the absence of prefixes can even lead to improved performance. This is because, without task-specific cues, the source prompts are encouraged to learn more general input similarities across tasks, allowing private prompts to specialize in capturing task-specific distinctions.

\subsection{Effect of Task Labels}

In prompt tuning, particularly for generative models such as T5-base, the semantics of the target labels can significantly influence the model's ability to generate correct outputs. To assess this, we measured the effect of label design on task clustering and knowledge transfer in our framework.

\begin{figure}
	\centering
	\includegraphics[width=0.7\linewidth]{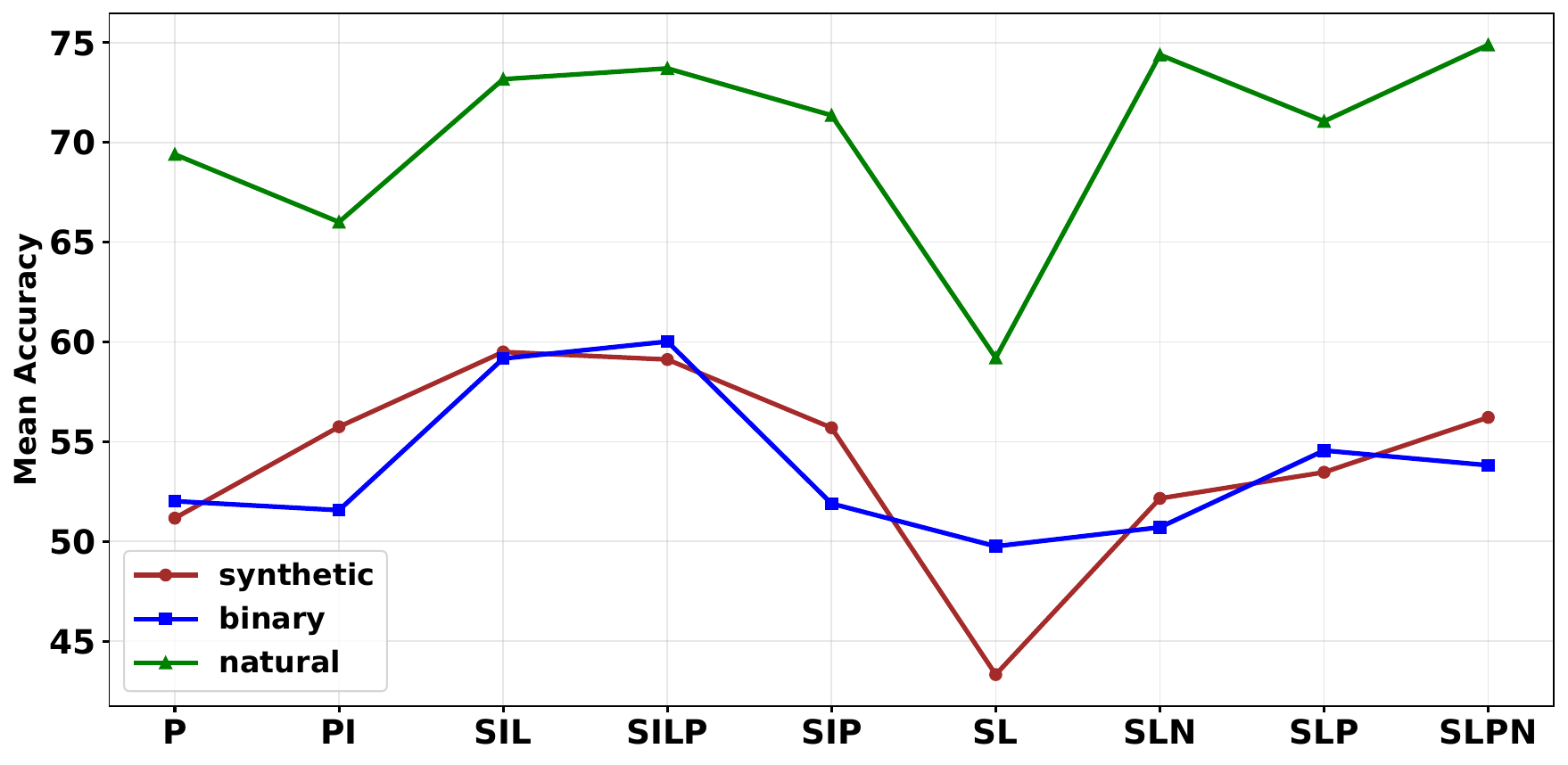}
	\caption{Effect of using natural, synthetic, and standardized numeric task labels on performance across GLUE tasks with 32 training samples. Semantically meaningful natural labels promote better knowledge transfer and clustering in prompt tuning, while synthetic and standardized numeric labels lead to interference and reduced performance.}
	\label{fig:labels}
\end{figure}

Figure~\ref{fig:labels} compares performance under three label settings: (1) natural, semantically meaningful class labels (e.g., \textit{entailment} vs. \textit{not entailment}), (2) synthetic, task-specific labels with no semantic overlap, and (3) standardized numeric labels shared across tasks (e.g., 0/1 for binary tasks, 0/1/2 for MNLI).

Natural labels often align across tasks (e.g., QNLI and RTE both use entailment-related labels) or are semantically close (such as \textit{equivalent} in MRPC and \textit{duplicate} in QQP), promoting shared representations in the model. In contrast, synthetic labels deliberately remove semantic connections by assigning arbitrary, task-specific symbols (e.g., A0, A1 for one task, B0, B1 for another), while standardized numeric labels create superficial label sharing even across unrelated tasks (e.g., SST-2 and QNLI both using 0/1).

As shown in Figure~\ref{fig:labels}, using natural labels consistently improves performance across all configurations. Both synthetic and standardized numeric labels result in lower performance, likely due to interference from unrelated tasks that share label indices without meaningful semantic overlap, lack of effective knowledge transfer between tasks, and limited alignment with representations learned during pretraining. These effects highlight the importance of semantically coherent label design for enabling robust prompt-based transfer. 

\subsection{Visualization of Attention Weights} \label{sec:sim}

\begin{figure}[h!]
	\centering
	\includegraphics[width=1\textwidth]{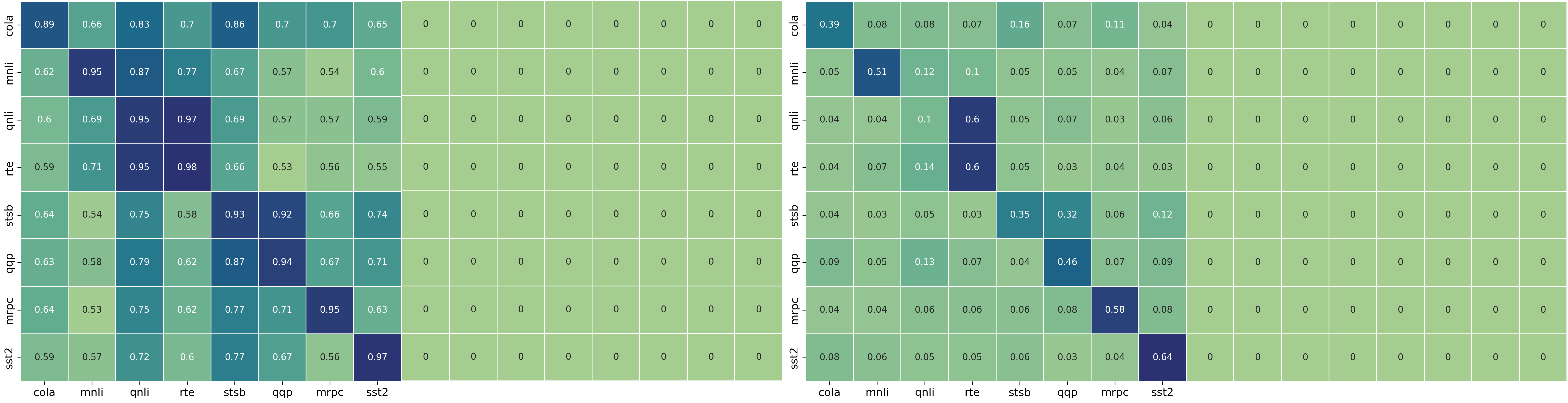}
	\includegraphics[width=1\textwidth]{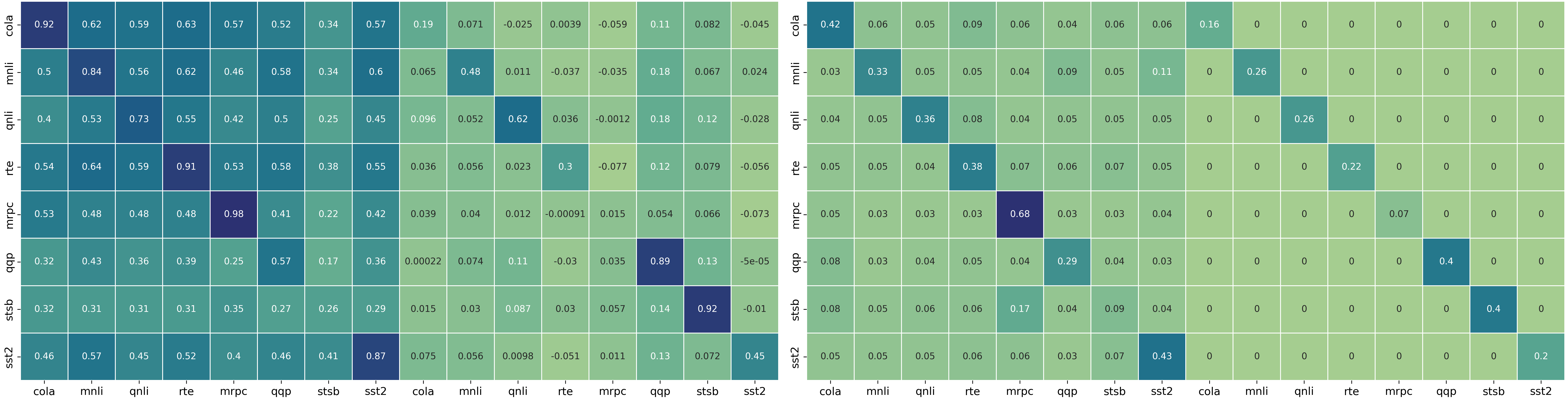}
	\includegraphics[width=1\textwidth]{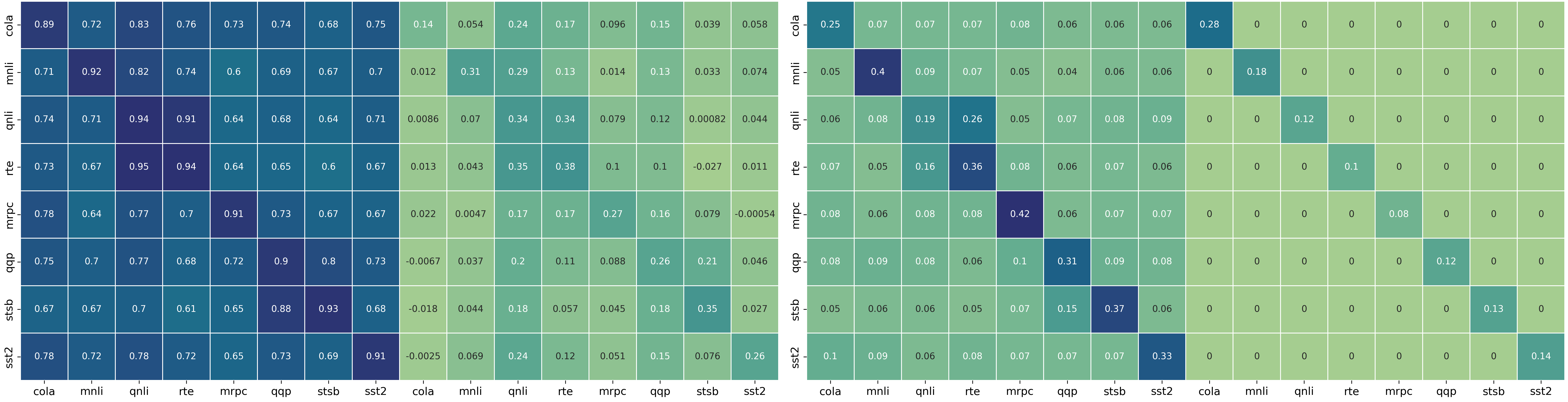}
	\includegraphics[width=1\textwidth]{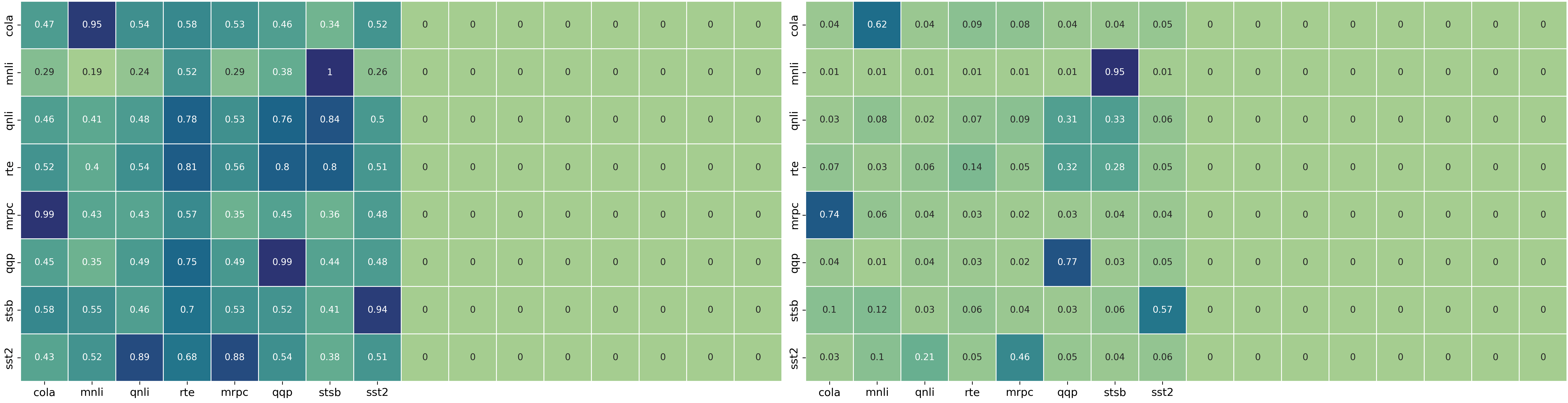}
	\includegraphics[width=1\textwidth]{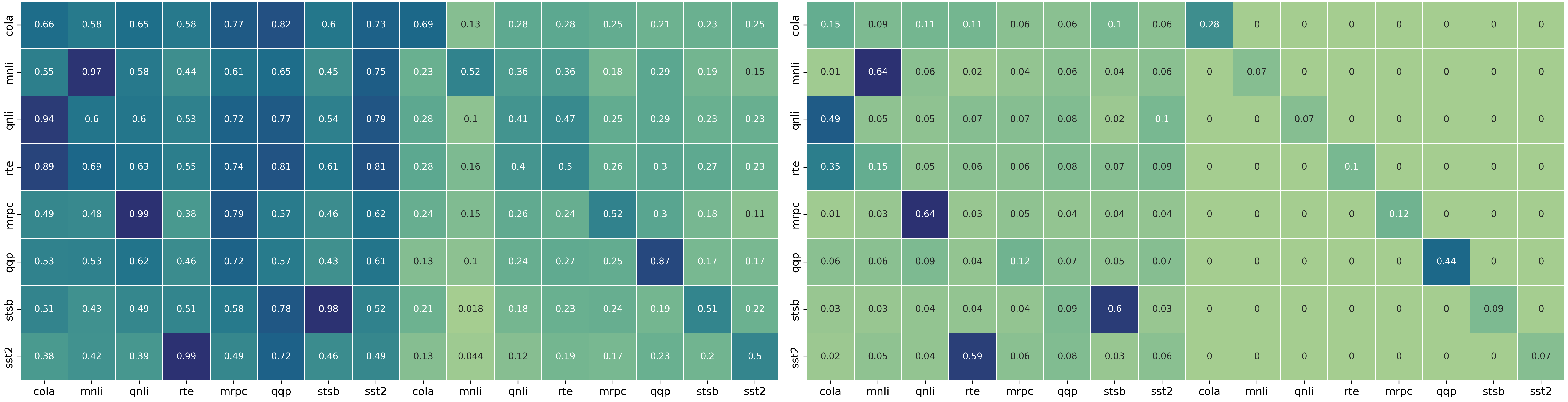}
	\caption[image]{Right: Heatmaps displaying attention scores for methods (top to bottom: SIL, SIP, SILP, SNL, and SLPN) trained on GLUE tasks with 32 samples. Left: Corresponding task similarities between the target prompt and each source and private prompt.}
	\label{fig:group_si}
\end{figure}

Figure~\ref{fig:group_si} visualizes the behavior of the attention module learned during target prompt training for different methods, using 32 training examples. 
The \textbf{right heatmaps} in these figures display the attention weights, which indicate the proportional contribution of each source prompt in constructing the target prompt, as formulated in Equation~\ref{eq:pt}. 
The \textbf{left heatmaps} show the cosine similarity between the target prompt and each source or private prompt, computed as:

\begin{align}
	\text{Sim}_{ij} = \cos(\mathbf{t}_i, \mathbf{s}_j) = 
	\frac{\mathbf{t}_i \cdot \mathbf{s}_j}{\|\mathbf{t}_i\| \, \|\mathbf{s}_j\|}
\end{align}

where \( \mathbf{t}_i \) is the mean embedding of the \( i \)-th target prompt, and \( \mathbf{s}_j \) is the mean embedding of the \( j \)-th source or private prompt. A higher similarity value suggests greater alignment or compatibility between the target and the corresponding source/private prompt.

Across SIL and SLN, where private prompts are absent, source prompts learn to specialize for individual tasks, with some prompts shared across similar tasks (e.g. QNLI source prompt). In these configurations, we observe that attention weights focus strongly on the most relevant source prompt, forming two main clusters: one for NLI tasks and another for the paraphrasing group, including STSB.

In SIP, SILP, and SLNP, private prompts play a more prominent role in capturing task-specific knowledge, allowing source prompts to generalize further. This is reflected in higher similarity measures between target and source prompts. In these cases, tasks with conflicting objectives or label ambiguity (e.g., STSB as regression, QQP overlapping with QNLI/MRPC) tend to rely more on private prompts. In SLNP, where all prompts are learned jointly but private prompts are available, we see a similar division of labor: source prompts remain broadly shared while private prompts specialize for difficult or conflicting tasks. Overall, the inclusion of private prompts consistently enhances flexibility in handling task similarities and differences within the framework.

\subsection{Source Prompt Similarity Metric} 

\begin{figure}
	\centering
	\includegraphics[width=0.5\linewidth]{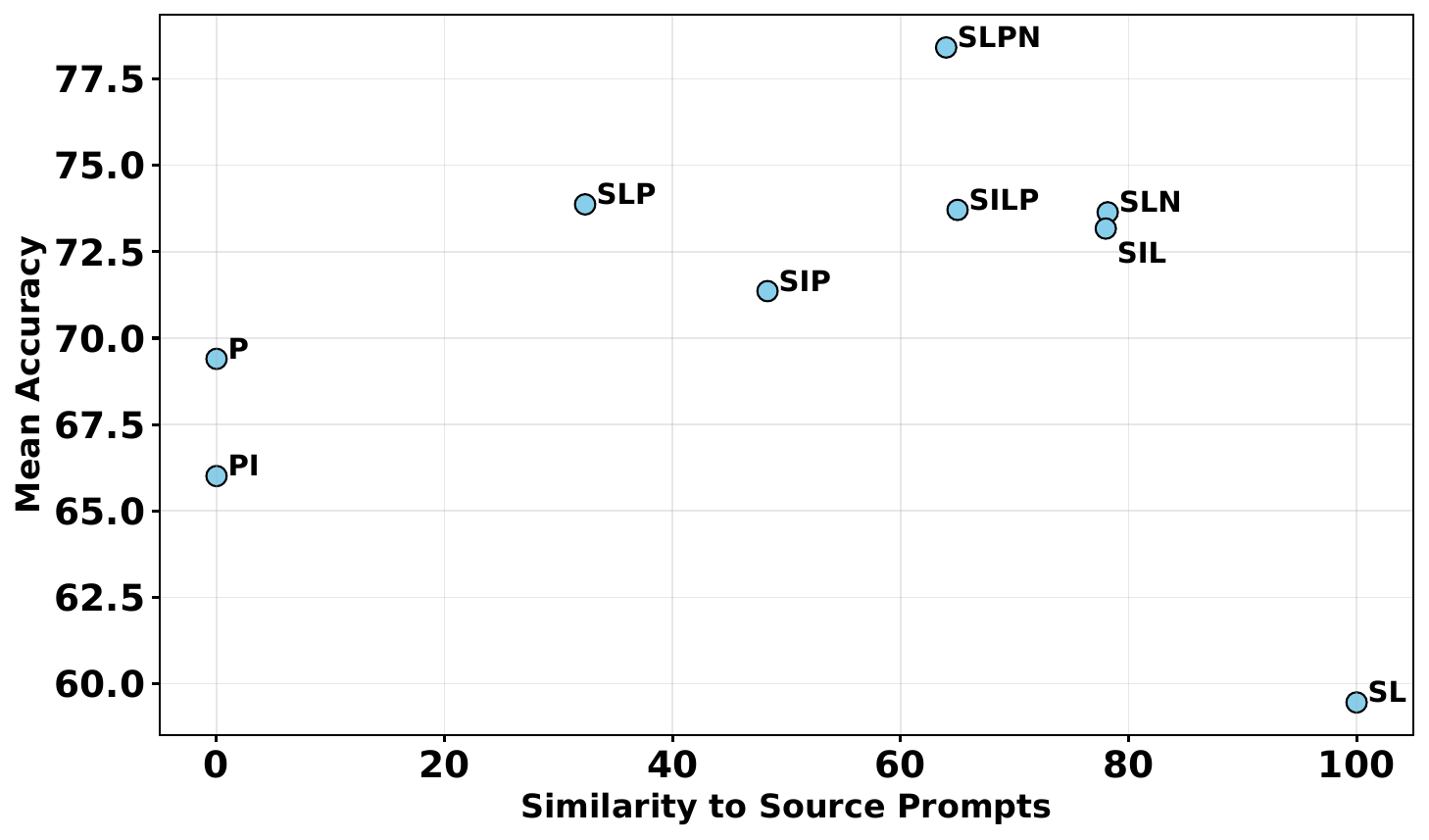}
\caption{Weighted Similarity to Source Prompts across different configurations on GLUE tasks with 32 training samples. Higher values indicate stronger reliance on shared source prompts.}

	\label{fig:wsim}
\end{figure}

To systematically quantify how much target prompts rely on source prompts across different methods, we introduce a metric called \emph{Weighted Similarity to Source Prompts}. This measure summarizes the attention-based similarity captured in the right panel of Figure~\ref{fig:group_si} presented above. Figure~\ref{fig:wsim} reports the values of this metric for the proposed methods. It is computed as the weighted average of attention scores between target and source prompts:

\begin{align}
	\text{WeightedSim}_{\text{source}} = 
	\frac{1}{N} \sum_{i=1}^{N} 
	\sum_{j \in \mathcal{S}} 
	w_{ij} \cdot \text{Sim}_{ij}
\end{align}

In this formulation, $\mathcal{S}$ denotes the set of source prompts, $w_{ij}$ is the attention weight assigned by target prompt $i$ to source prompt $j$, and $\text{Sim}_{ij}$ represents their similarity. This metric reflects the extent to which target prompts leverage shared knowledge encoded in the source prompts.

As expected, the SL configuration, which uses a single source prompt shared across all tasks, achieves the maximum value (100) for this metric, yet shows the lowest performance overall. In contrast, P and PI have no source prompts by design, so the metric is undefined for them. SIL and SLN exhibit similar metric values and comparable performance, reflecting shared reliance on source prompts with moderate specialization. SILP and SLNP also show similar metric values, but SLNP achieves higher performance on GLUE tasks due to joint learning from scratch, which promotes better information exchange. 

Finally, SLP and SIP yield relatively lower metric values: in SLP the single source prompt limits sharing capacity, while in SIP frozen source prompts restrict adaptation. These patterns suggest that achieving good performance requires a balanced degree of source similarity---neither too low, which limits knowledge transfer, nor too high, which risks overgeneralization and failure to capture task-specific distinctions.

\subsection{Number of Source Prompts}  
\begin{figure}[h!]
	\centering
	\includegraphics[width=0.7\textwidth]{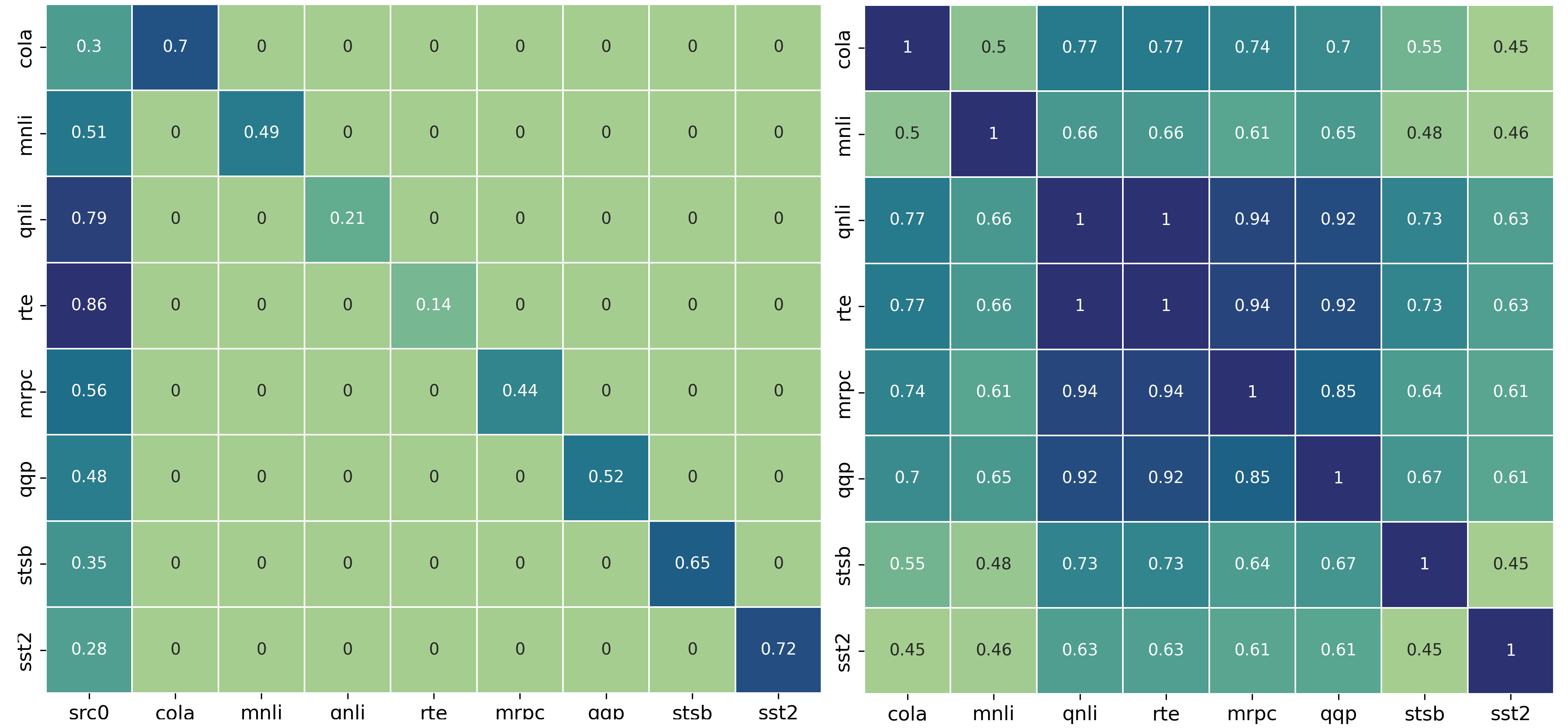}
	\includegraphics[width=0.7\textwidth]{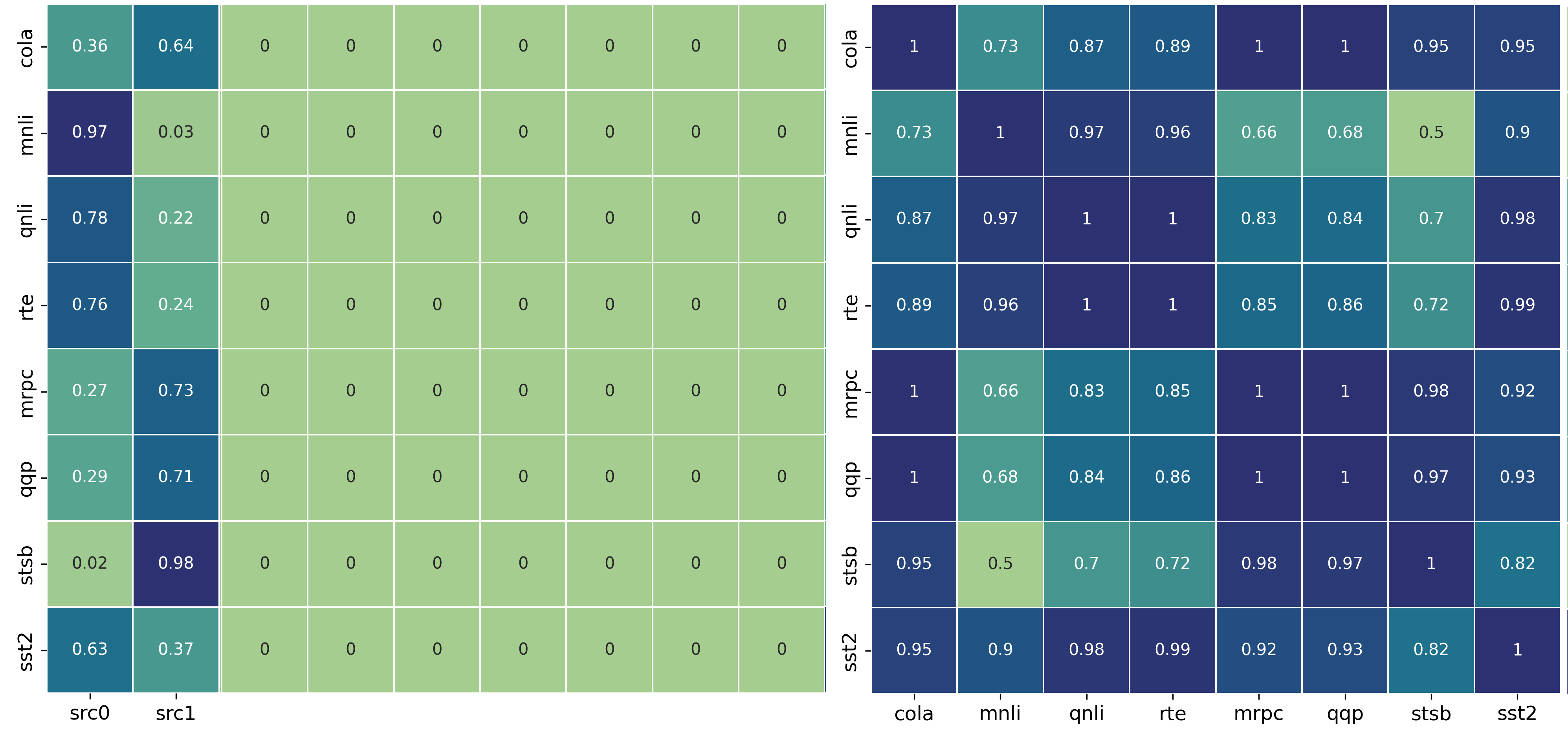}
	
	\caption[image]{ On the left, heatmaps illustrate attention scores for the SLP (top) and SL2 (bottom) methods, trained on GLUE tasks with 32 samples. On the right, the corresponding cross-task similarity measures are shown.
	}
	\label{fig:group_slp}
\end{figure}

Since source prompts are intended to capture general knowledge across tasks, we chose to experiment with a smaller number of source prompts. Figure~\ref{fig:group_slp} visualizes SLP with a single shared source prompt and SL with two source prompts (referred to as SL2). 

We further measured cross-task target prompt similarity using normalized token-level cosine similarity:

\begin{align}
	S_{ij} = \frac{ \displaystyle \frac{1}{mn} \sum_{k=1}^{m} \sum_{l=1}^{n} \cos(\mathbf{t}_{ik}, \mathbf{t}_{jl}) }{ \sqrt{ S_{ii} \cdot S_{jj} } }
\end{align}

This is shown in the right panel of Figure~\ref{fig:group_slp} for each setting. In the SLP case, attention weights and task similarity indicate that the single source prompt is primarily shared by analogous tasks (QNLI, RTE, MNLI, MRPC, QQP), while dissimilar tasks (COLA, STSB, SST2) rely more on private prompts. MNLI, with its more diverse label set, attends to both shared and private prompts.  

As shown in the rightmost panel of Figure~\ref{fig:group_slp}, even though STSB and MNLI rely more on private prompts, they still exhibit similarity with other tasks. This is expected, as sentence similarity (STSB) and causal reasoning (MNLI) can support paraphrasing and NLI tasks. In SL with two source prompts, tasks are more clearly separated across these prompts, with one primarily covering NLI tasks and the other focusing on paraphrasing tasks.

\begin{figure}
	\centering
	\includegraphics[width=0.5\linewidth]{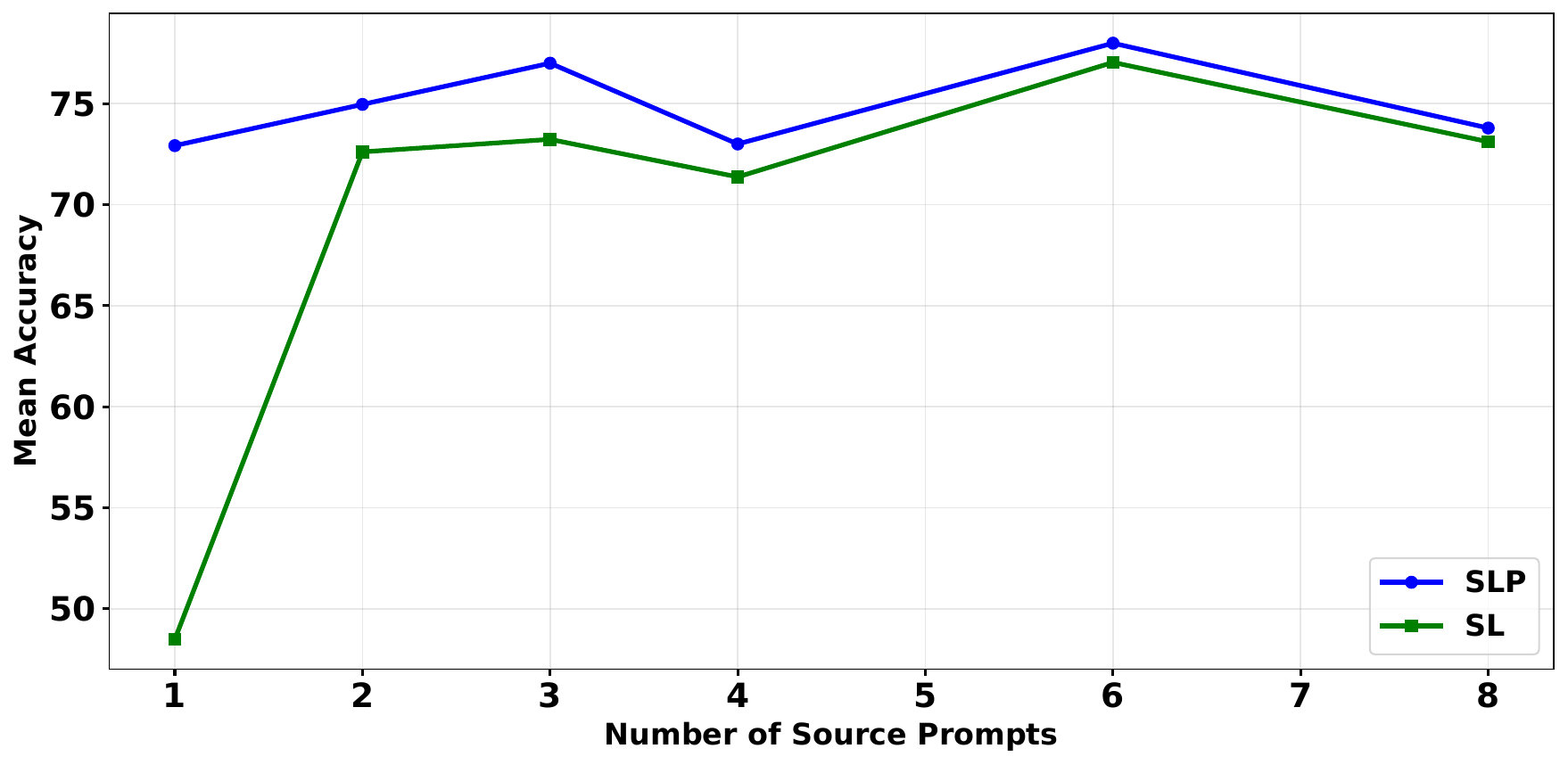}
\caption{Performance on GLUE tasks with 32 training samples as a function of the number of source prompts for SL and SLP configurations. Results show that increasing the number of source prompts substantially improves SL performance, while SLP remains more stable due to the presence of private prompts.}
	\label{fig:nsp}
\end{figure}

For SL and SLP, we also examined the effect of increasing the number of source prompts on performance. Figure~\ref{fig:nsp} shows GLUE task performance across different numbers of source prompts for both settings. We observe a sharp performance increase for SL when moving from one to two source prompts, as this provides more room for tasks to differentiate while still allowing them to share knowledge within clusters of related tasks. The resulting task clustering and similarity matrix for two prompts is depicted in Figure~\ref{fig:group_slp}. 

However, increasing the number of source prompts beyond a certain point can lead to diminishing returns or even degraded performance. When too many prompts are used, they may become overly fragmented, reducing the capacity to concentrate shared knowledge across tasks. This can result in overparameterization and a higher risk of undertraining, especially in low-resource scenarios where available data or training epochs are insufficient to effectively optimize all prompts. In our experiments, using around three source prompts typically yielded strong performance on GLUE, aligning well with natural task clusters. While increasing to eight prompts improved capacity in some cases, it also led to slight declines in performance, highlighting the need to balance representation power with learning stability.

\subsection{Balancing the Influence of Source and Private Prompts}

\begin{figure}
	\centering
	\includegraphics[width=0.5\linewidth]{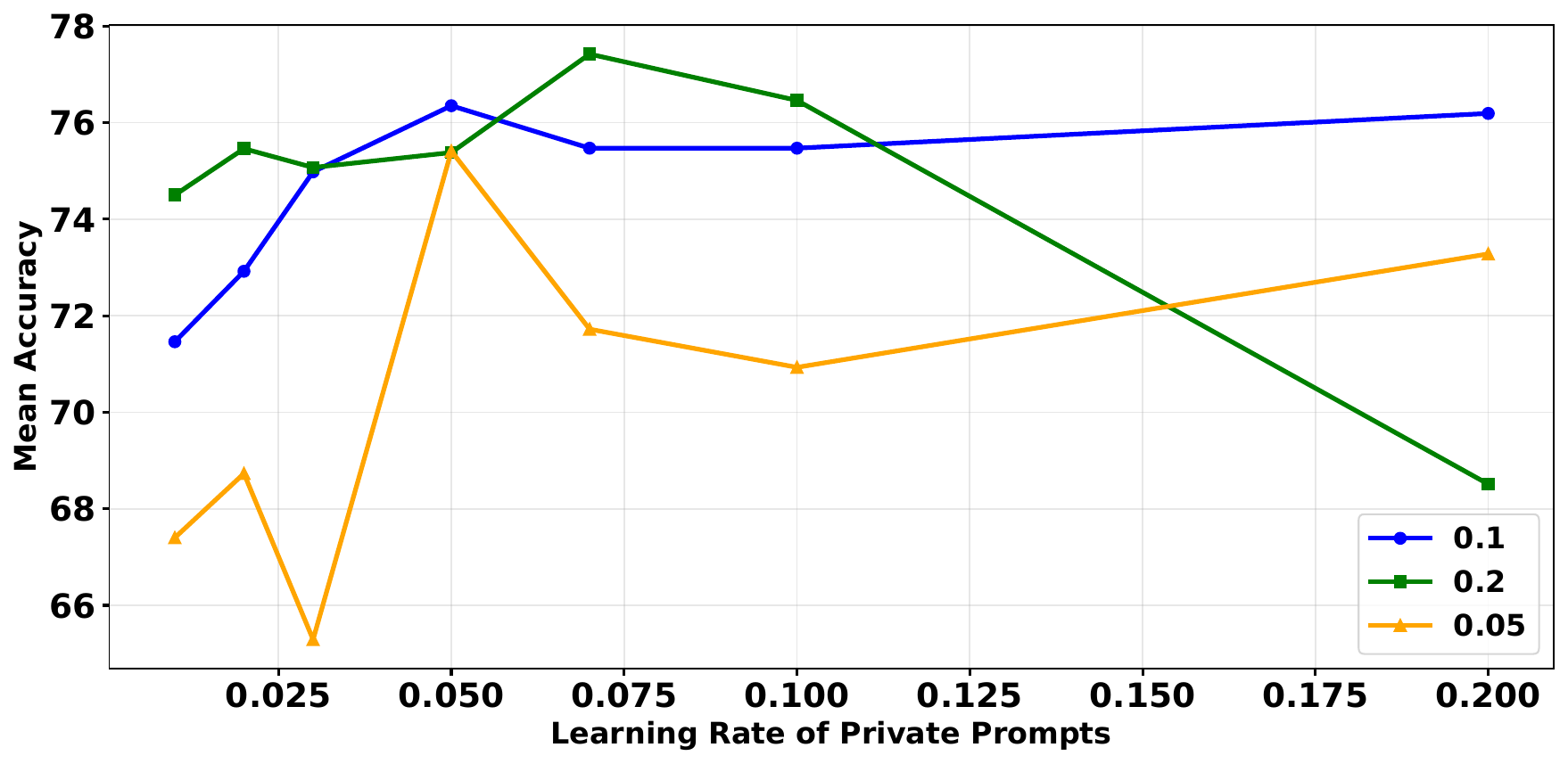}
\caption{Effect of varying source and private prompt learning rates on overall GLUE performance of SLP method (32 samples per task). Higher learning rates for source prompts improve generalization by promoting shared knowledge across tasks. Optimal performance is typically achieved when the private prompt learning rate is approximately half of the source prompt rate, enabling fine-grained task-specific adaptation without destabilizing shared representations.}

	\label{fig:lr}
\end{figure}

Beyond controlling the number of source prompts, another way to regulate the contribution of source versus private prompts is through their respective learning rates. We evaluated the effect of varying source and private prompt learning rates on overall performance of SLP setting, as shown in Figure~\ref{fig:lr}.

The results indicate that higher learning rates for source prompts generally lead to improved performance, particularly when the private prompt learning rate is kept relatively lower. The optimal configuration typically occurs when the learning rate for private prompts is approximately half that of the source prompts. For instance, when the source prompt learning rate is set to 0.2, increasing the private prompt learning rate to match or exceed it leads to a sharp performance drop. In contrast, a more moderate source learning rate (e.g., 0.1) yields more stable performance across a range of private learning rates.

This behavior can be explained by the functional roles of the prompts. Source prompts are responsible for capturing generalizable knowledge across tasks, and therefore benefit from a more aggressive learning rate that facilitates broader pattern acquisition. In contrast, private prompts are intended to specialize in task-specific nuances. A lower learning rate in this case allows for finer-grained adaptation without overwhelming shared representations. Balancing these learning rates is thus essential to achieving effective multi-task prompt tuning, ensuring that source prompts generalize while private prompts personalize.

\section{Related Works}
This work belongs to the line of parameter-efficient fine-tuning (PEFT) for pre-trained language models \cite{peft}. PEFT approaches aim to adapt large pre-trained models to downstream tasks while updating only a small subset of parameters. Instead of modifying all weights as in standard fine-tuning, they introduce compact trainable components such as adapters, bias terms, soft prompts, or low-rank matrices. This enables efficient adaptation with reduced memory and compute costs.

\paragraph{Adapter-based methods.} Houlsby et al. \cite{houlsby2019parameter} introduced adapters, small neural modules inserted between Transformer layers. We evaluate this in the full-data setting as \textbf{Adapter(m)}. BitFit \cite{zaken2022bias}, included as \textbf{BitFit}, fine-tunes only the bias terms. HyperFormer and HyperDecoder \cite{hyper}, shown as \textbf{HyperFormer(m)} and \textbf{HyperDecoder(m)}, are adapter-based variants with additional parameterization for improved flexibility. 

\paragraph{Prompt tuning and its variants.} Vanilla Prompt Tuning (PT) \cite{lester2021power, gptund, mto}, included in both our full and few-shot tables as \textbf{PT}, learns continuous prompts prepended to inputs.

\paragraph{Decomposed Prompt Tuning.} Several works aim to reduce the cost and improve the flexibility of soft prompt tuning by factorizing or decomposing the prompt \cite{dept, dept2, compt}. DEPT \cite{dept} proposes Decomposed Prompt Tuning, which splits a soft prompt into a shorter prompt plus low-rank matrices optimized separately. This design reduces memory and time costs while maintaining or improving performance over vanilla prompt tuning and its variants. We include it as \textbf{DEPT} in both the full-data and few-shot comparisons.

\paragraph{Prompt transfer and multi-task approaches.} Vu et al. \cite{spot} introduced SPoT (\textbf{SPoT} in our tables), which improves prompt tuning across tasks via soft prompt transfer. It offers multi-task and source-prompt selection variants.  Shen et al. \cite{mvpt} further extended this paradigm to vision?language models. Wang et al. \cite{rel-multitask} proposed Multitask Prompt Tuning (MPT, shown as \textbf{MPT} and \textbf{MPT(m)}), which learns a single transferable prompt via multi-task learning with prompt decomposition and distillation.

\paragraph{Attention-based prompt mixing.} Asai et al. \cite{rel-attempt} proposed ATTEMPT (\textbf{ATTEMPT} and \textbf{ATTEMPT(m)}), which learns to interpolate between source and private prompts via an instance-wise attention module. Unlike their approach, which freezes source prompts during target training, our method pre-trains and adapts source prompts to each target task.

\begin{table*}[ht!]
	\centering
	\caption{Comparison with related works on GLUE tasks (full-data setting). Accuracy is reported for all tasks except STS-B (Pearson correlation). (m) indicates the multi-tasking variant of each method.}

	\label{table:related}
	\begin{tabular}{lccccccccl}
		\toprule
		\textbf{method/ task} & \textit{mnli} & \textit{qqp} & \textit{qnli} & \textit{sst2} & \textit{stsb} & \textit{mrpc} & \textit{rte} & \textit{cola} & \textbf{avg.} \\
		\toprule
		
		\textbf{FineTuning} & 86.8 & 91.6 & 93.0 & 94.6 & 89.7 & 90.2 & 71.9 & 61.8 & 84.9 \\
		
		\textbf{Fine-tuning(m)} & 85.7 & 91.1 & 92.0 & 92.5 & 88.8 & 90.2 & 75.4 & 54.9 & 83.8 \\
		
		\textbf{BitFit} & 85.3 & 90.1 & 93.0 & 94.2 & 90.9 & 86.8 & 67.6 & 58.2 & 83.3 \\
		
		\textbf{Adapter(m)} & 86.3 & 90.5 & 93.2 & 93.0 & 89.9 & 90.2 & 70.3 & 61.5 & 84.4 \\
		
		\textbf{HyperFormer(m)} & 85.7 & 90.0 & 93.0 & 94.0 & 89.7 & 87.2 & 75.4 & 63.7 & 84.8 \\
		
		\textbf{HyperDecoder(m)} & 86.0 & 90.5 & 93.4 & 94.0 & 90.5 & 87.7 & 71.7 & 55.9 & 83.7 \\
		
		\textbf{SPoT} & 85.4 & 90.1 & 93.0 & 93.4 & 90.0 & 79.7 & 69.8 & 57.1 & 82.3 \\
		
		\textbf{ATTEMPT} & 84.3 & 90.3 & 93.0 & 93.2 & 89.7 & 85.7 & 73.4 & 57.4 & 83.4 \\

		\textbf{ATTEMPT(m)} & 83.8 & 90.0 & 93.1 & 93.7 & 90.8 & 86.1 & 79.9 & 64.3 & 85.2 \\
		
		\textbf{MPT} & 85.9 & 90.3 & 93.1 & 93.8 & 90.4 & 89.1 & 79.4 & 62.4 & 85.6 \\

		\textbf{MPT(m)} & 84.3 & 90.0 & 93.0 & 93.3 & 90.4 & 89.2 & 82.7 & 63.5 & 85.8 \\
		
		\textbf{DEPT} & 85.0 & 90.4 & 93.2 & 94.2 & 90.8 & 90.7 & 79.1 & 63.8 & 85.9 \\
		
		\textbf{SLP10} (ours) & 82.1 & 89.6 & 92.8 & 93.2 & 89.3 & 86.0 & 73.3 & 81.8 & \textbf{86.0} \\
		
		\bottomrule
	\end{tabular}
\end{table*}

\begin{table*}[h!]
	\centering
\caption{Comparison on GLUE tasks in few-shot setting ($k=16$ and $k=32$ samples per class).}

	\label{table:k16_32}
	\begin{tabular}{lcccccccccl}
		\toprule
		\textbf{method/ task} & \textit{mnli} & \textit{qqp} & \textit{qnli} & \textit{sst2} & \textit{stsb} & \textit{mrpc} & \textit{rte} & \textit{cola} & \textbf{avg.} \\
		\midrule
		\multicolumn{10}{c}{\textbf{16 samples}} \\
		\midrule
		\textbf{PT}    & 41.5 & 62.3 & 59.9 & 50.9 & 87.8 & 68.1 & 54.7 & 28.5 & 56.7 \\
		\textbf{MPT}   & 61.6 & 84.7 & 90.6 & 63.2 & 89.1 & 70.1 & 64.8 & 32.1 & 69.5 \\
		\textbf{DEPT}  & 61.8 & 80.3 & 91.2 & 77.6 & 87.1 & 78.1 & 71.9 & 27.1 & 71.9 \\
		\textbf{SLPN} (ours) & 55.67 & 79.00 & 84.00 & 87.77 & 85.40 & 61.67 & 60.67 & 61.67 & \textbf{71.98} \\
		\midrule
		\multicolumn{10}{c}{\textbf{32 samples}} \\
		\midrule
		\textbf{PT}    & 37.0 & 62.3 & 56.7 & 50.9 & 87.5 & 68.1 & 54.7 & 23.2 & 55.1 \\
		\textbf{MPT}   & 63.6 & 88.5 & 91.0 & 75.9 & 89.7 & 74.5 & 59.7 & 30.8 & 71.7 \\
		\textbf{DEPT}  & 63.3 & 80.1 & 91.3 & 80.4 & 89.2 & 81.4 & 72.7 & 28.6 & 73.4 \\
		\textbf{SLPN} (ours) & 66.25 & 76.25 & 84.25 & 91.35 & 86.52 & 75.75 & 54.25 & 74.25 & \textbf{76.11} \\
		\bottomrule
	\end{tabular}
\end{table*}

\subsection{Comparison with Related Works} In contrast to these approaches, our method (\textbf{SLP}, \textbf{SLPN}) directly leverages multiple source prompts during target prompt training. Unlike prior work, we focus on generative language models where label semantics play a crucial role in distilling pre-trained knowledge---particularly in prompt tuning settings---and where related tasks can effectively reinforce each other through shared representations. Unlike ATTEMPT, which calculates attention scores by comparing the target task and input, we learn an attention distribution over multiple source prompts. This design enables richer knowledge sharing among tasks and more effective adaptation. 

As shown in Table~\ref{table:related}, our approach slightly outperforms related methods in the full-data setting. The SLP10 configuration uses 10 source prompts---approximately two more than the number of tasks---demonstrating that with ample data, model capacity can be increased by employing more source prompts to better capture shared knowledge. More importantly, Table~\ref{table:k16_32} shows that our method achieves noticeably stronger results in few-shot scenarios, particularly with 32 training examples per task---a low-resource setting where our approach is able to learn effectively from scratch while still leveraging shared representations.

\section{Discussion}

\subsection{Division of Labour}
CrossPT introduces a multi-task prompt tuning framework that explicitly divides the representational burden between shared \emph{source prompts} and task-specific \emph{private prompts}. This design enables the source prompts to capture generalizable, transferable knowledge across related tasks, while private prompts adapt this information to task-specific nuances. Our results demonstrate that relying exclusively on private prompts (vanilla prompt tuning) or only on shared prompts is suboptimal, confirming the benefit of combining both components.

\subsection{Transfer from High-Resource to Low-Resource Tasks}
This separation of shared and private components is particularly effective in transfer scenarios with imbalanced data availability. For example, in the QQP--MRPC experiments (Section~\ref{sec:mrpc-qqp}), the low-resource MRPC task benefited significantly from representations shaped by the richer QQP data. Such results highlight CrossPT's ability to leverage high-resource tasks to improve performance on low-resource, challenging tasks.

\subsection{Balancing Source and Private Prompts}
A crucial design consideration in CrossPT is determining the number of shared source prompts. While increasing $S$ can enhance the model's ability to cluster tasks and capture broader patterns (e.g., distinguishing NLI from paraphrasing tasks), excessive fragmentation can harm performance due to reduced shared capacity and increased computational cost. Our experiments suggest there is an optimal range for $S$ that balances generalization and efficiency.

Additionally, learning-rate control is essential to balance the roles of source and private prompts. Assigning higher learning rates to source prompts encourages them to quickly capture shared structures, while using lower rates for private prompts allows stable adaptation to task-specific details. In practice, setting the private prompt learning rate to approximately half that of the source prompts provided robust results.

\subsection{Computational Efficiency}  
CrossPT is designed for parameter-efficient adaptation. Let $k$ be the prompt length and $d$ the embedding dimension. Each target task has its own private prompt, so with $S$ shared source prompts and $T$ target tasks, the total number of trainable parameters is:
\[
(S + T) \times k \times d + d^2 + d,
\]
accounting for the prompt token embeddings and the shared linear prompt encoder (weights $\mathbf{W} \in \mathbb{R}^{d \times d}$ and bias $\mathbf{b} \in \mathbb{R}^d$). This is substantially smaller than the full model parameter count in standard fine-tuning.

Training cost grows linearly with $(S + T)$, as all source and private prompt embeddings are updated. At inference time, CrossPT uses all $S$ source prompts along with the single private prompt corresponding to the target task. This leads to an inference complexity of:
\[
O(S \times k \times d^2),
\]
which remains lightweight compared to full fine-tuning.

Moreover, the model's capacity can be flexibly adjusted based on data availability by varying the number of source prompts. For example, in our experiments, the best performance in the full-data setting was achieved using 10 source prompts, illustrating how additional capacity can improve results when sufficient data is available.

\subsection{Modularity and Flexibility}
CrossPT retains a modular design, supporting flexible choices of source and target tasks. While in our experiments we often used the same task set for both roles, this joint training improved overall performance by encouraging cross-task learning. However, CrossPT can easily incorporate additional source tasks to improve transfer. For example, in Group 2 experiments, adding QNLI as a source task has been shown to be beneficial across multiple targets (see Section~\ref{sec:sim}). Conversely, excluding less relevant source tasks (e.g., CoLA in GLUE) can reduce unnecessary complexity without hurting performance. Because each task has its own private prompt, tasks with less shared structure naturally rely more on their private prompts, allowing the model to handle heterogeneous task sets without enforcing unnecessary sharing.

\subsection{Limitations and Future Work}

Despite its advantages, CrossPT introduces certain trade-offs. Selecting the optimal number and composition of source prompts requires careful tuning and may not generalize across domains or tasks. Moreover, although the model significantly reduces the number of updated parameters compared to full fine-tuning, memory and computation costs still increase with the number of source prompts, limiting scalability.

Several directions remain for future work. First, more adaptive mechanisms for source prompt selection, such as dynamic pruning or relevance-based filtering, could mitigate computational overhead. Second, enhancing generalization to tasks with open-ended outputs---such as question answering---and extending the framework to multi-modal settings like vision-language models would broaden the applicability of CrossPT. Third, improving robustness to variations in label semantics and task formulation is essential for real-world deployment. Finally, leveraging instance-specific information (e.g., input features or uncertainty) during prompt composition could enable more fine-grained adaptation and improve performance on diverse tasks.

\section{Conclusion}

In this study, we proposed Cross-task Prompt Tuning (CrossPT), a modular framework designed to improve multi-task learning by combining shared source prompts with task-specific private prompts. Our approach enables parameter-efficient adaptation of large language models while promoting effective cross-task knowledge transfer.

Our experiments on the GLUE benchmark and related tasks show that carefully balancing shared and private components---via prompt initialization, adaptive attention mechanisms, and tuning the number of source prompts---substantially improves performance. Specifically, CrossPT achieves up to 3---5 percentage points higher accuracy than conventional single-task prompt tuning in few-shot settings with 32 examples per task, while maintaining strong parameter efficiency.

We also demonstrated that design choices such as incorporating task prefixes and semantically meaningful target labels further enhance transferability and generalization. Additionally, our analysis introduced new metrics, such as Weighted Similarity to Source Prompts, to quantify knowledge sharing and specialization across tasks.

Overall, our findings highlight that prompt-based adaptation benefits significantly from deliberate control over shared and private information. By selecting appropriate configurations, practitioners can achieve competitive performance with minimal additional parameters while reducing task interference. In conclusion, CrossPT provides a practical and versatile approach for scalable, efficient, and effective multi-task prompt tuning, laying the groundwork for future research in cross-task transfer and low-resource learning.

\bibliography{sn-bibliography}

\begin{appendices}
\section{Task Performance Summary}
\subsection{Results for GLUE tasks}
Below, we present a detailed overview of the average performance achieved on selected tasks within the GLUE dataset:
\begin{table}[htbp]
	\centering
	\begin{tabular}{lccccccccc}
		\toprule
		\textbf{Conf} & \textbf{All} & \textbf{CoLA} & \textbf{MNLI} & \textbf{MRPC} & \textbf{QNLI} & \textbf{QQP} & \textbf{RTE} & \textbf{SST-2} & \textbf{STS-B} \\
		\midrule
		SLPN & 71.98 $\pm$ 4.12 & 61.67 & 55.67 & 61.67 & 79.00 & 84.00 & \textbf{60.67} & 87.77 & \textbf{85.40} \\
		SILP & 71.65 $\pm$ 1.08 & 53.67 & \textbf{62.00} & 61.33 & 85.33 & 85.33 & 54.00 & 88.23 & 83.33 \\
		SLP  & 71.19 $\pm$ 2.11 & \textbf{67.00} & 55.33 & 67.33 & 66.67 & 83.33 & 54.67 & \textbf{90.87} & 84.27 \\
		SIL  & 70.81 $\pm$ 2.41 & 45.00 & 60.33 & 60.67 & 86.67 & 85.67 & 54.00 & 89.00 & 85.13 \\
		SLN  & 69.91 $\pm$ 3.36 & 55.00 & 50.33 & \textbf{68.33} & 80.67 & 75.67 & 55.33 & 89.43 & 84.50 \\
		SIP  & 62.38 $\pm$ 7.06 & 34.00 & 59.67 & 63.00 & \textbf{88.33} & 83.33 & 58.00 & 83.30 & 29.37 \\
		SL   & 59.67 $\pm$ 0.62 & 31.33 & 45.00 & 56.67 & 77.67 & \textbf{86.67} & 53.00 & 47.33 & 79.70 \\
		PI   & 59.23 $\pm$ 1.54 & 49.33 & 52.33 & 58.00 & 87.33 & 83.00 & 55.33 & 85.43 & 3.10 \\
		P    & 56.37 $\pm$ 3.60 & 39.33 & 44.67 & 50.00 & 76.67 & 80.67 & 50.33 & 64.33 & 44.97 \\
		
		\bottomrule
	\end{tabular}
	\caption{Performance (mean $\pm$ std) of different prompt configurations on 16-sample GLUE tasks. Best values per task are in bold.}
	\label{tab:mets}
\end{table}

\begin{table}[htbp]
	\centering
	\begin{tabular}{lccccccccc}
		\toprule
		\textbf{Conf} & \textbf{All} & \textbf{CoLA} & \textbf{MNLI} & \textbf{MRPC} & \textbf{QNLI} & \textbf{QQP} & \textbf{RTE} & \textbf{SST-2} & \textbf{STS-B} \\
		\midrule
		SLPN & 76.11 $\pm$ 4.78 & \textbf{74.25} & \textbf{66.25} & \textbf{75.75} & 76.25 & 84.25 & 54.25 & \textbf{91.35} & \textbf{86.52} \\
		SLP  & 74.32 $\pm$ 3.64 & 70.50 & 58.75 & 64.75 & \textbf{86.75} & 83.50 & 57.50 & 88.68 & 84.15 \\
		SILP & 73.71 $\pm$ 2.12 & 57.67 & 63.00 & 66.33 & 83.67 & 87.00 & 56.33 & 90.87 & 84.80 \\
		SIL  & 73.17 $\pm$ 1.86 & 53.33 & 60.67 & 66.33 & 85.00 & 87.33 & 56.33 & 90.43 & 85.87 \\
		SLN  & 73.15 $\pm$ 3.56 & 66.29 & 59.43 & 68.14 & 85.71 & 85.43 & 55.29 & 87.14 & 77.80 \\
		SIP  & 71.36 $\pm$ 0.92 & 47.00 & 64.00 & 60.00 & 84.33 & 87.67 & 57.00 & 86.10 & 84.77 \\
		P    & 69.40 $\pm$ 1.96 & 62.00 & 60.67 & 56.00 & 84.00 & 72.00 & 52.00 & 82.30 & 86.20 \\
		PI   & 66.01 $\pm$ 5.12 & 40.67 & 58.00 & 60.00 & 86.00 & \textbf{87.67} & \textbf{60.33} & 82.67 & 52.73 \\
		SL   & 53.59 $\pm$ 18.74 & 37.29 & 41.43 & 53.43 & 70.29 & 61.14 & 50.86 & 61.91 & 52.37 \\
		
		\bottomrule
	\end{tabular}
	\caption{Performance (mean $\pm$ std) of different prompt configurations on 32-sample GLUE tasks. Best values per task are in bold.}
	\label{tab:mets32}
\end{table}

\subsection{Results for Group 2 tasks}
Below, we present a detailed overview of the average performance achieved on selected tasks within the Group 2:

\begin{table}[htbp]
	\centering
	\begin{tabular}{lcccccccc}
		\toprule
		\textbf{Conf} & \textbf{All} & \textbf{IMDB} & \textbf{MRPC} & \textbf{MultiNLI} & \textbf{PAWS} & \textbf{SciTail} & \textbf{STSB2} & \textbf{Yelp} \\
		\midrule
		SILP & 68.35 $\pm$ 2.37 & 77.80 & 72.00 & 63.67 & \textbf{56.70} & \textbf{72.23} & 48.43 & 87.57 \\
		SLPN & 68.30 $\pm$ 2.31 & 71.10 & 73.67 & 61.67 & 53.00 & 64.00 & \textbf{71.57} & 83.13 \\
		SLP  & 67.56 $\pm$ 1.66 & 74.13 & \textbf{78.67} & 60.90 & 50.33 & 58.67 & 62.10 & 88.13 \\
		SIP  & 65.37 $\pm$ 3.20 & 75.90 & 67.67 & \textbf{64.91} & 48.00 & 69.70 & 43.53 & 88.77 \\
		SLN  & 64.54 $\pm$ 3.42 & 75.20 & 70.67 & 62.43 & 51.10 & 57.23 & 45.33 & \textbf{89.80} \\
		PI   & 63.92 $\pm$ 0.72 & \textbf{81.23} & 63.33 & 61.10 & 50.80 & 69.00 & 39.20 & 82.77 \\
		SL   & 62.92 $\pm$ 2.98 & 75.47 & 63.00 & 62.77 & 50.57 & 59.10 & 42.57 & 87.00 \\
		SIL  & 62.64 $\pm$ 0.61 & 72.43 & 56.33 & 64.90 & 47.23 & 69.67 & 41.43 & 86.47 \\
		P    & 59.19 $\pm$ 2.27 & 56.10 & 72.33 & 58.00 & 47.10 & 60.10 & 34.23 & 86.43 \\
		
		\bottomrule
	\end{tabular}
	\caption{Performance (mean $\pm$ std ) of prompt configurations on additional tasks (Group 2) using 32 samples. Bolded values are the best per task.}
	\label{tab:mets2}
\end{table}

\end{appendices}

\end{document}